\documentclass[final,3p,times]{elsarticle}

\usepackage{zmy_template}

\journal{}
\begin{document}
\begin{frontmatter}

\title{Transferring self-supervised pre-trained models for SHM data anomaly detection with scarce labeled data}

\author[1]{Mingyuan Zhou~\corref{c1}}
\author[1,2,3]{Xudong Jian~\corref{c1}}
\author[4,5]{Ye Xia}
\author[1,6]{Zhilu Lai}
\cortext[c1]{Corresponding authors. E-mail address: xudong.jian@sec.ethz.ch (Xudong Jian) \& mzhou151@connect.hkust-gz.edu.cn (Mingyuan Zhou)}

\affiliation[1]{organization={The Internet of Things Thrust, The Hong Kong University of Science and Technology (Guangzhou)},
            state={Guangzhou},
            country={China}}

\affiliation[2]{organization={Future Resilient Systems, Singapore-ETH Centre},
            state={Singapore},
            country={Singapore}}

\affiliation[3]{organization={Department of Civil, Environmental and Geomatic Engineering, ETH Zurich},
            state={Zurich},
            country={Switzerland}}

\affiliation[4]{organization={Department of Bridge Engineering, Tongji University},
            state={Shanghai},
            country={China}}            

\affiliation[5]{organization={Shanghai Qi Zhi Institute},
            state={Shanghai},
            country={China}}    

\affiliation[6]{organization={The Department of Civil and Environmental Engineering, The Hong Kong University of Science and Technology },
            state={Hong Kong},
            country={China}} 

\begin{abstract}
Structural health monitoring (SHM) has experienced significant advancements in recent decades, accumulating massive monitoring data. 
Data anomalies inevitably exist in monitoring data, posing significant challenges to their effective utilization.
Recently, deep learning has emerged as an efficient and effective approach for anomaly detection in bridge SHM. 
Despite its progress, many deep learning models require large amounts of labeled data for training.
The process of labeling data, however, is labor-intensive, time-consuming, and often impractical for large-scale SHM datasets. 
To address these challenges, this work explores the use of self-supervised learning (SSL), an emerging paradigm that combines unsupervised pre-training and supervised fine-tuning.
The SSL-based framework aims to learn from only a very small quantity of labeled data by fine-tuning, while making the best use of the vast amount of unlabeled SHM data by pre-training. 
Mainstream SSL methods are compared and validated on the SHM data of two in-service bridges.
Comparative analysis demonstrates that SSL techniques boost data anomaly detection performance, achieving increased $F_1$ scores compared to conventional supervised training, especially given a very limited amount of labeled data. 
This work manifests the effectiveness and superiority of SSL techniques on large-scale SHM data, providing an efficient tool for preliminary anomaly detection with scarce label information.
\end{abstract}



\end{frontmatter}


\section{Introduction}
Structural health monitoring (SHM) has experienced noticeable progress due to the compulsory requirement of structural safety and the rapid development of sensor technology~\cite{ko2005technology,avci2021review,sun2020review}.
The primary objective of SHM is to assess the health and integrity of structures by acquiring and analyzing various types of data, such as acceleration, strain, etc.
As time evolves and the number of installed sensors grows, SHM systems accumulate increasing volumes of monitoring data. 
These data hold essential information for evaluating structural conditions and are leveraged in a variety of SHM tasks. Typical tasks include structural damage identification~\cite{lai2019semi,an2019recent}, structural system identification~\cite{lai2021structural,green2015bayesian}, load identification~\cite{jian2022robust,wang2021static}, and structural life cycle management~\cite{frangopol2019life,chen2023innovative}, etc. 

Data-driven SHM tasks rely on the quantity and quality of the acquired data. However, data anomalies inevitably exist and pose significant challenges to the performance and reliability of SHM approaches. 
For example, abnormal data may severely impact the accuracy of vibration model estimation, particularly in high-order modes~\cite{deng2024abnormal}.
Abnormal data can also distort signal processing, resulting in erroneous power spectral density calculations and false alarms in structural damage identification~\cite{fu2019sensor,kullaa2011distinguishing}.
In addition, SHM data anomalies exhibit diverse patterns and may arise from various sources, including sensor malfunctions, environmental changes, or structural damages. 
Therefore, detecting and classifying these data anomalies is critical for ensuring the effective functioning of data-driven SHM approaches. 

Data anomaly detection is widely applied across diverse research and application domains~\cite{chandola2009anomaly}. 
In the field of SHM, Deng et al.~\cite{deng2024abnormal} present a comprehensive and up-to-date review of approaches for SHM data anomaly detection.
These methods are broadly classified into categories such as statistical probability methods, predictive models, computer vision methods, etc.
Many statistical probability and predictive methods require prior information on data distribution, statistical model parameters, manually set thresholds, or expertise inspection.
However, applying these approaches to SHM data can be challenging due to the tremendous volume and intricate patterns.

To address these challenges, researchers have developed reliable and automatic approaches for SHM data anomaly detection. 
Deep learning techniques have been recently extensively employed, demonstrating significant effectiveness in classifying various abnormal patterns in SHM data~\cite{bao2019computer,tang2019convolutional,jian2021faulty,lei2023mutual,ni2020deep}.
A common workflow involves transforming time series data into images and then training deep learning models.
For instance, Bao et al.~\cite{bao2019computer} employ a convolutional neural network (CNN) to detect bridge abnormal data. Lei et al.~\cite{lei2023mutual} utilize the attention neural network, showing improved classification accuracy and computational efficiency.
Despite the success of deep learning methods, they often rely heavily on large labeled datasets for supervised learning.
However, creating large labeled datasets requires expert knowledge and considerable labor, presenting an obstacle to performing data anomaly detection more efficiently.

To reduce the reliance on labeled data, researchers have developed approaches that require less to no labeled data, categorized as \textit{unsupervised} or \textit{semi-supervised} methods.
Unsupervised methods~\cite{mao2021toward,entezami2023continuous,sarmadi2023unsupervised}, which do not need labeled data, typically train neural networks to reconstruct the input data. After training, reconstruction errors are utilized as indicators to detect anomalies.
While unsupervised methods perform well in distinguishing between normal and abnormal data, they generally cannot classify multiple abnormal patterns. 
In contrast, semi-supervised methods~\cite{pan2023transfer,zhang2022data_2,wang2023data} effectively classify multiple abnormal patterns by leveraging a small amount of labeled data. 
These approaches come with specific \textit{prerequisites}.
For instance, some methods require training data that contains solely normal data~\cite{mao2021toward}, others rely on data augmentation for specific types of anomalies~\cite{wang2023data}, and some utilize pre-trained models from other labeled datasets~\cite{pan2023transfer}. These requirements can be challenging in practical SHM applications. 
Therefore, developing more flexible data anomaly methods with minimal prerequisites remains a challenging and open problem.

To develop the approach with less labeled data and unknown data distribution, we introduce self-supervised learning (SSL) for SHM data anomaly detection.  
SSL~\cite{balestriero2023cookbook,gui2024survey,he2020momentum,chen2020simple,misra2020self} have recently emerged as a powerful approach for learning data representations without the need for manually annotated labels.
So far, SSL has become predominant in various domains, from large language models such as GPTs~\cite{brown2020language}, BERT~\cite{devlin2018bert}, to advanced image models like SEER~\cite{goyal2021self}, their pre-training is employed in a self-supervised manner.
Unlike supervised learning, SSL requires no manually annotated labels during pre-training, enabling it to learn from vast amounts of unlabeled data.
The key idea of SSL is to design pretext tasks that leverage data itself or its augmentation as label information. Typical pretext tasks include reconstruction and comparison, which allow models to learn useful representations for downstream tasks~\cite{chen2021exploring,liu2021self}. 
A typical SSL workflow is to leverage vast unlabeled data for \textit{pre-training}, followed by supervised \textit{fine-tuning}~\cite{chen2020big}. 
Despite its success in many domains, the integration of SSL pre-training and fine-tuning for SHM data anomaly detection remains underexplored.
Given the large scale of unlabeled SHM data, employing SSL on SHM data becomes a promising solution.

Our work is geared toward learning from a few labeled SHM data while making the best use of a vast amount of unlabeled SHM data for anomaly detection. The key contributions of this study are as follows:
\begin{itemize}
    \item We introduce emerging self-supervised learning techniques in the context of SHM data anomaly detection. The SSL-based framework performs pre-training on SHM data without relying on prior knowledge of data distributions. Using the data itself as supervision, SSL eliminates the need for manually annotated labels during the pre-training stage.
    \item Considering large-scale data and high cost of labeling in SHM applications, we fine-tune the pre-trained SSL model using only a small amount of (a few hundred) labeled samples.
    This makes the method highly practical for real-world applications where labeled data is scarce and expensive to obtain.
    \item We conduct a comparative analysis of four mainstream SSL methods and conventional supervised training on various abnormal patterns across two bridges. The results validate the effectiveness of SSL pre-training and provide a guideline for employing SSL techniques for SHM data anomaly detection.
\end{itemize}

\label{}

\section{Methodology}
\begin{figure*}[!htb]
    \centering
    \includegraphics[width=\linewidth]{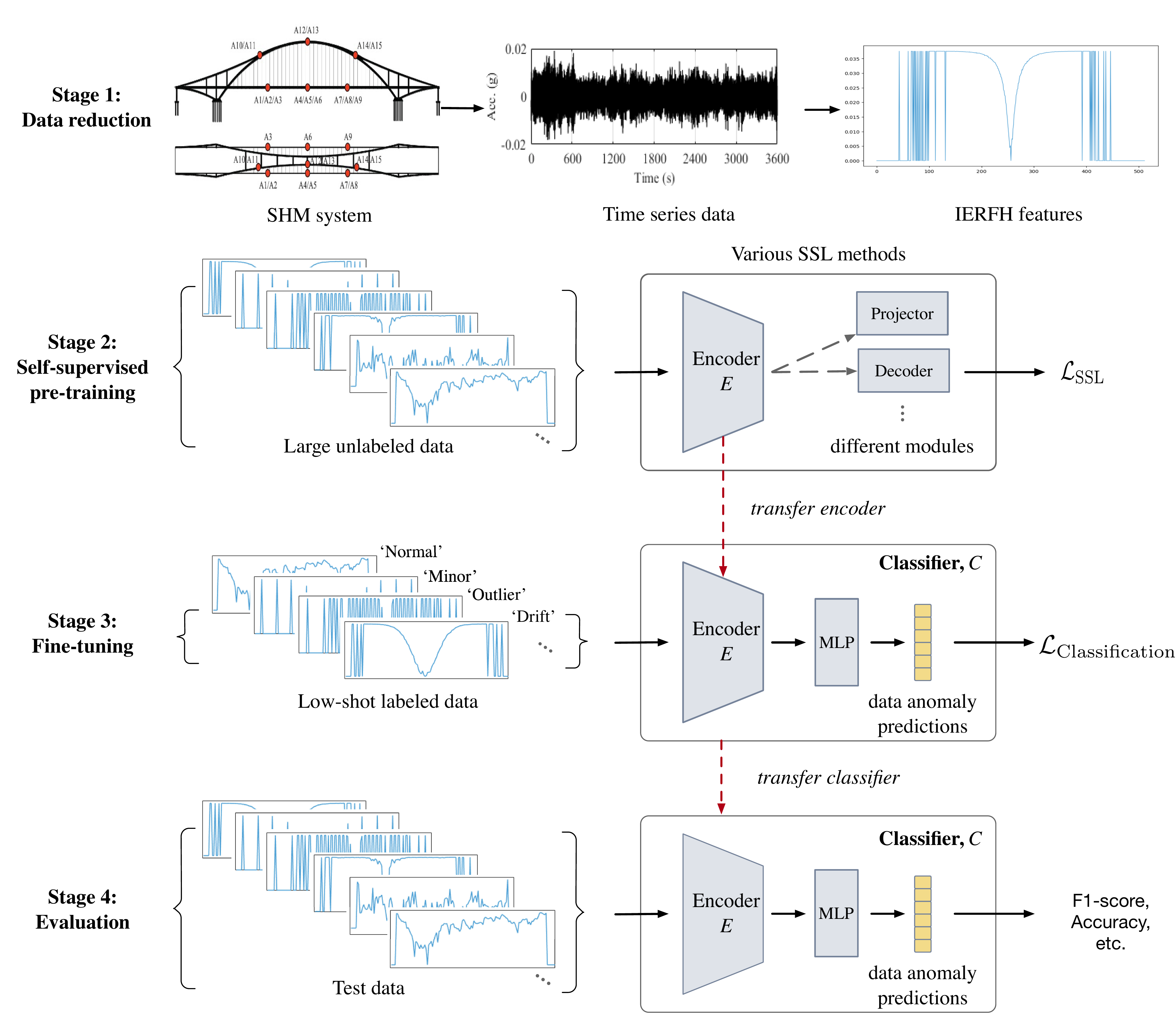}
    \caption{The workflow of the self-supervised learning framework for data anomaly detection. (Details about the IERFH feature are provided in Section \ref{sec:data_reduction}). }
    \label{fig:ssl_framework}
\end{figure*}

\subsection{Problem formulation}
Our work addresses the potential challenge of label scarcity in SHM data anomaly detection, where only a small fraction of data is labeled in a much larger unlabeled dataset.
In particular, we consider a large unlabeled dataset, $\mathcal{D} = \{\mathbf{x}_i\}_{i=1}^{N}$, comprising $N$ data samples $\mathbf{x}_i$, and a \textit{low-shot} dataset, $\mathcal{D}_{\text{low-shot}} = \{\mathbf{x}_i, y_i\}_{i=1}^{M}$, containing $M$ labeled samples and corresponding labels $y$, where $M \ll N$.
The labeled dataset $\mathcal{D}_{\text{low-shot}}$ is a subset of the unlabeled dataset $\mathcal{D}$, i.e., $\mathcal{D}_{\text{low-shot}} \subset \mathcal{D}$.
The SSL-based framework involves first pre-training a model via self-supervised learning on the large unlabeled dataset $\mathcal{D}$, and then applying supervised fine-tuning using the small labeled dataset $\mathcal{D}_{\text{low-shot}}$.
After fine-tuning, the trained model can detect anomalies in SHM applications or be evaluated on test datasets.

The overall workflow of the SSL-based framework is illustrated in Figure~\ref{fig:ssl_framework}.
In the first stage, high-dimensional time series SHM data are transformed into low-dimensional features, specifically the inverted envelope of its relative frequency histogram (IERFH)~\cite{devore1995probability,jian2021faulty}.
Secondly, self-supervised learning is employed to pre-train a neural network using the IERFH features extracted from the large unlabeled SHM dataset. Subsequently, the pre-trained neural network is fine-tuned on the classification task using the small labeled dataset.
After fine-tuning, the model is evaluated on the test dataset to validate its performance on data anomaly detection.

\subsection{Data reduction}
\label{sec:data_reduction}
As shown in stage 1 of Figure~\ref{fig:ssl_framework}, data reduction is a common and necessary step for deep learning when handling high-dimensional observations.
SHM systems continuously monitor structures, generating time series data that are often too high-dimensional to be directly used as inputs for deep learning models.
Therefore, data reduction is performed as the preliminary step before pre-training.
In our implementation, we reduce the dimension of measured acceleration data by extracting the inverted envelope of its relative frequency histogram (IERFH).

A relative frequency distribution demonstrates how frequently each unique value appears within a set of discrete values, and it is typically represented using a histogram.
Particularly, the relative frequency distribution histogram (RFDH) describes a group of discrete data in the probability domain, effectively representing the original time series data~\cite{devore1995probability}.
For extracting RFHD, we follow standard practices where the left and right edges of the horizontal axis correspond to the measurement range of the sensors.
To ensure a smooth and representative envelope while maintaining computational efficiency, the number of bins for a 1-hour segment of SHM data is set to 512 in this study.
Additionally, the histogram is normalized to display relative frequencies, with the upper and lower limits of the vertical axis being 1 and 0, respectively.

The IERFH has proven to be an effective feature for data anomaly detection in supervised settings~\cite{jian2021faulty}.
In our SSL framework, we utilize the IERFH to reduce the dimension of SHM time series data.
Specifically, 1-hour segments of SHM time series data are transformed into their IERFH features.
These features serve as the input data, denoted as $\mathbf{x}$, for deep learning models, where $\mathbf{x} \in \mathbb{R}^{512}$.
By converting high-dimensional time-series data into a compact, 512-dimensional feature space, the data reduction stage provides a computationally efficient and informative input for training deep learning models.

\subsection{Self-supervised pre-training}
\label{sec:pretraining}
\begin{figure*}[!htb]
    \begin{subfigure}{\textwidth}
        \centering
        \includegraphics[width=14cm]{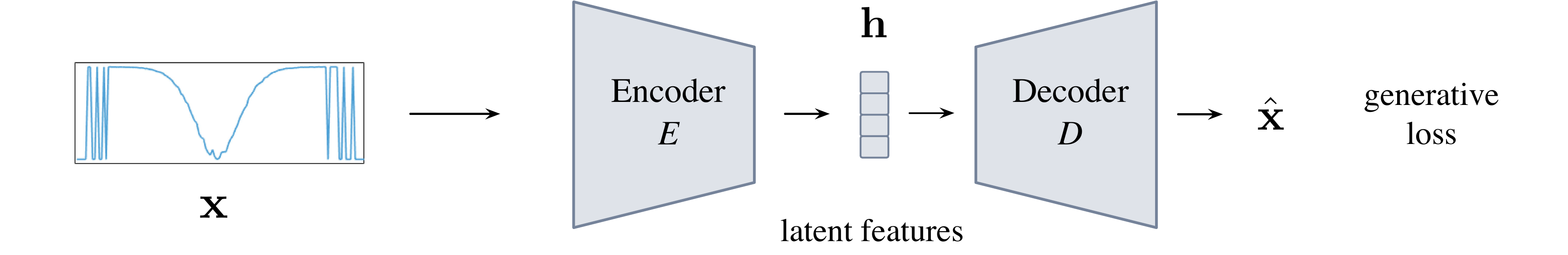}
        \caption{\textit{Generative} pretext task}
        \label{fig:ssl_ae}
    \end{subfigure}
    \begin{subfigure}{\textwidth}
        \centering
        \includegraphics[width=14cm]{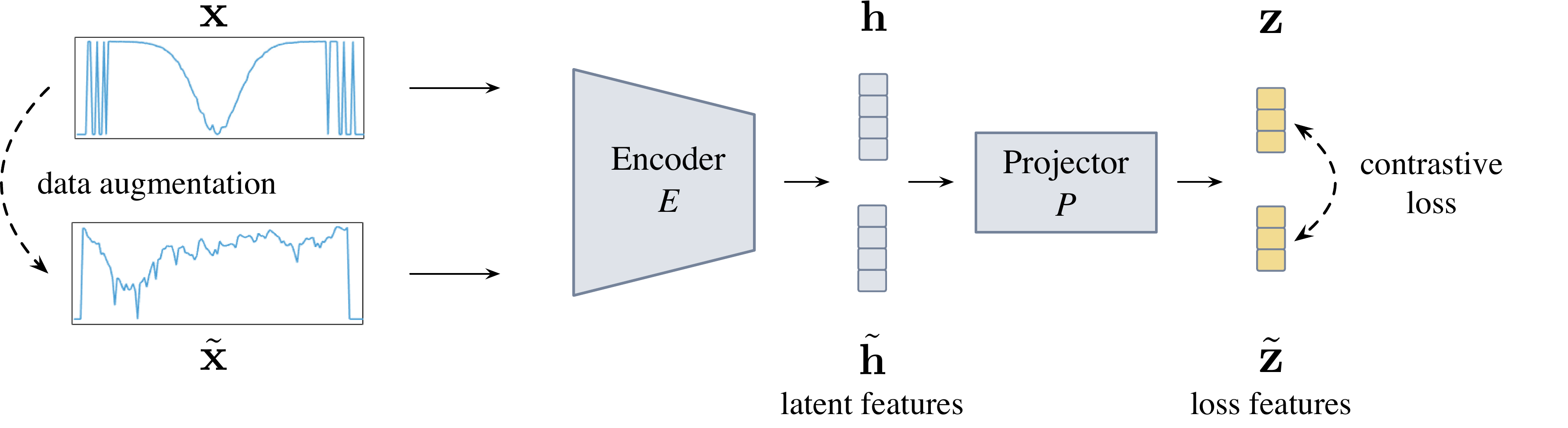}
        \caption{\textit{Contrastive} pretext task}
        \label{fig:ssl_contrast}
    \end{subfigure}
    \begin{subfigure}{\textwidth}
        \centering
        \includegraphics[width=14cm]{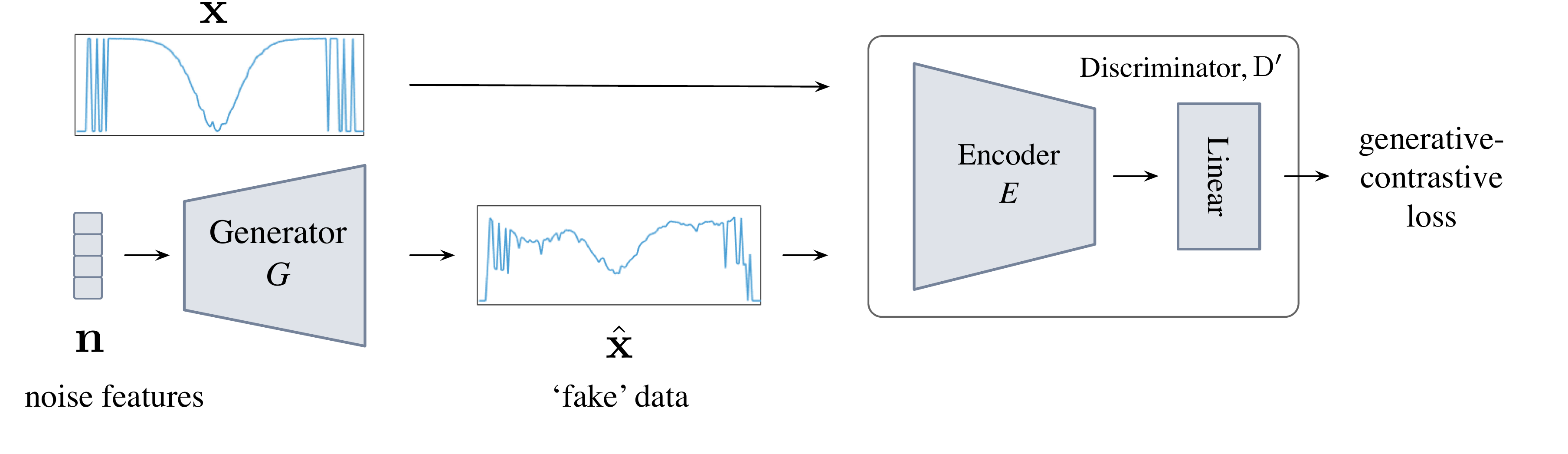}
        \caption{\textit{Generative-contrastive} pretext task}
        \label{fig:ssl_gan}
    \end{subfigure}
    \caption{Categories of pretext tasks in self-supervised learning: different pretext tasks aim to train an encoder for learning feature representations without requiring manually annotated labels.}
    \label{fig:ssl_details}
\end{figure*}

The purpose of self-supervised pre-training is to learn general and expressive data representations by utilizing the data itself or its augmented versions as label information.
As illustrated in stage 2 in Figure~\ref{fig:ssl_framework}, the pre-training involves training an encoder $E$ to extract representations from data, which can be readily adapted to downstream tasks, such as data anomaly detection (classification).
In particular, self-supervised pre-training is guided by pretext tasks designed to encourage the model to understand inherent data patterns.
These tasks typically involve reconstructing input data or comparing different data samples, often with the assistance of additional modules like projectors or decoders.
SSL pretext tasks can be categorized into three types~\cite{liu2021self}: \textit{generative}, \textit{contrastive} and \textit{generative-contrastive}.
Figure~\ref{fig:ssl_details} provides an overview of different pretext tasks used in SSL, and the following sections introduce each category in detail. 

\subsubsection{Generative self-supervised learning}
In \textit{generative} SSL, the pretext task aims to train neural networks to reconstruct the original input data through an encoder-decoder architecture.
As presented in Figure~\ref{fig:ssl_ae}, the encoder $E$ is designed to encode input data $\mathbf{x}$ into latent feature representation $\mathbf{h}$, while the decoder $D$ is utilized to generate reconstruction $\hat{\mathbf{x}}$ from $\mathbf{h}$.
From a feature-learning perspective, the decoder is able to reconstruct data samples from the latent features, indicating the encoder can learn representative patterns of the input data.
The generative reconstruction process is as follows:
\begin{equation}
    \hat{\mathbf{x}} = D(E(\mathbf{x})).
\end{equation}
Typical generative SSL models include auto-encoders (AE), flow models~\cite{dinh2016density}, auto-regressive models~\cite{yang2019xlnet}, etc.
For instance, the generative loss function of an AE is defined as:
\begin{equation}
    \mathcal{L}_{\text{SSL-AE}} = \frac{1}{B} \sum_{i=1}^{B} ||\hat{\mathbf{x}}_i - \mathbf{x}_i||_2^2 \ ,
    \label{eq:ae}
\end{equation}
where $B$ denotes the batch size and $\mathcal{L}_{\text{SSL-AE}}$ represents the mean-squared error (MSE) between the reconstructed data $\hat{\mathbf{x}}$ and the original input data $\mathbf{x}$.
In the implementation, we utilize a convolutional neural network (CNN)-based encoder.
CNNs are suitable for finding local patterns and preserving spatial hierarchies within data~\cite{lecun1998gradient,krizhevsky2012imagenet}.
Unless otherwise specified, this CNN-based encoder will be used throughout the remaining sections, with its architecture details provided in Section~\ref{sec:baseline}.

\subsubsection{Contrastive self-supervised learning}
\textit{Contrastive} SSL focuses on learning representations by encouraging similar data samples (positive pairs) to be ``close" in feature space while pushing dissimilar samples (negative pairs) ``far apart"~\cite{chen2020simple,wickstrom2022mixing}.
Therefore, contrastive pretext tasks enable neural networks to discover discriminative patterns in data.
As illustrated in Figure~\ref{fig:ssl_contrast}, contrastive SSL generally comprises four parts: a data augmentation operation, an encoder for learning general representations, a projector that maps general representations to the loss function space, and a contrastive loss function.
Below, we describe these parts in detail and then introduce two widely used contrastive SSL methods: SimCLR and Mixup.

Data augmentation aims to generate augmented views of a given data to encourage the model to learn similar representations.
For instance, adding Gaussian noise~\cite{chen2020simple} or mixing data samples~\cite{zhang2017mixup} creates augmented views of the input data.
Subsequently, the encoder $E$ is employed to learn data representations. 
Particularly, the encoder takes a data $\mathbf{x}$ as input and transforms it into latent features $\mathbf{h}$, expressed as:
\begin{equation}
    \mathbf{h} = E(\mathbf{x}),
\end{equation}
where $\mathbf{h} \in \mathbb{R}^{256}$ in our implementation.
Next, a projector $P$, typically a linear layer, then maps latent features $\mathbf{h}$ into a lower-dimensional space for contrastive loss computation:
\begin{equation}
    \mathbf{z} = P(\mathbf{h}) =P(E(\mathbf{x})),
\end{equation}
where $\mathbf{z} \in \mathbb{R}^{128}$ and $\mathbf{z}$ are the features in the loss function space.
Most contrastive SSL methods include a projector after the encoder, showing significant classification accuracy gains in applications~\cite{balestriero2023cookbook}.
In addition, the projector enhances the efficiency of the learning process by reducing the features' dimensions.
For training neural networks, contrastive loss functions compare data samples using a similarity function, such as cosine similarity.
For vectors $\mathbf{u}$ and $\mathbf{v}$, the cosine similarity is defined as:
\begin{align}
    \text{sim}(\mathbf{u}, \mathbf{v}) = \frac{\mathbf{u}^{\top}\mathbf{v}}{ \|\mathbf{u}\| \cdot \|\mathbf{v}\| }  , 
    \label{eq:cosine}
\end{align}
which $||\cdot||$ denotes the $L_2$-norm, and $\text{sim}(\mathbf{u}, \mathbf{v})$ is also the dot product between $L_2$ normalized $\mathbf{u}$ and $\mathbf{v}$.

\textbf{A simple framework for contrastive learning (SimCLR)}~\cite{chen2020simple} is a state-of-the-art contrastive SSL framework for learning visual representations.
The data augmentation in SimCLR involves creating two augmented views of original data.
For a data sample $\mathbf{x}_i$, augmentation transformations $\mathcal{T}_1$ and $\mathcal{T}_2$ generates two augmented views:
\begin{equation}
    \tilde{\mathbf{x}}^{(1)}_i = \mathcal{T}_1 (\mathbf{x}_i) \ , \ \tilde{\mathbf{x}}^{(2)}_i = \mathcal{T}_2 (\mathbf{x}_i),
\end{equation}
where $ \tilde{\mathbf{x}}^{(1)}_i$ and $ \tilde{\mathbf{x}}^{(2)}_i$ represent two augmented data samples.
The contrastive loss function for SimCLR is defined as:
\begin{align}
    \ell_{\text{SimCLR}}(\tilde{\mathbf{z}}^{(1)}_i, \tilde{\mathbf{z}}^{(2)}_i) &= -\log \frac{\exp \left(\text{sim}(\tilde{\mathbf{z}}^{(1)}_i, \tilde{\mathbf{z}}^{(2)}_i) / \tau\right)}{\sum_{k=1}^{2 B} \mathbbm{1}_{[k \neq i]} \exp \left(\text{sim}(\tilde{\mathbf{z}}^{(1)}_i, \mathbf{z}_{k}) / \tau\right)}, \\
    \mathcal{L}_{\text{SSL-SimCLR}} &= \frac{1}{2B} \sum_{i=1}^{B} \left[\ell_{\text{SimCLR}}(\tilde{\mathbf{z}}^{(1)}_i, \tilde{\mathbf{z}}^{(2)}_i) + \ell_{\text{SimCLR}}(\tilde{\mathbf{z}}^{(2)}_i, \tilde{\mathbf{z}}^{(1)}_i) \right],
    \label{eq:ssl_simclr}
\end{align}
where $B$ is the batch size, $\tau$ is a temperature parameter for adjusting the value of feature similarity, and $\mathbbm{1}_{[k \neq i]}$ is an indicator function to exclude self-similarity.
The contrastive loss in SimCLR aims to maximize the similarity between two augmented samples of the same data sample (positive pairs) while minimizing the similarity between augmented samples and other randomly sampled data samples (negative pairs).
    
\textbf{Mixing up contrastive learning (Mixup)} is another typical contrastive SSL framework, which has shown effectiveness on representation learning for image and time series~\cite{wickstrom2022mixing,zhu2023patch,kim2020mixco}.
Mixup operation, originally proposed by Zhang et al.~\cite{zhang2017mixup}, creates new data samples through convex combinations of two data samples.
Given two data samples $\mathbf{x}_i^{(1)}$ and $\mathbf{x}_i^{(2)}$, a new augmented sample is constructed as:
\begin{align}
   & \tilde{\mathbf{x}}_i = \lambda \mathbf{x}_i^{(1)} + (1-\lambda) \mathbf{x}_i^{(2)} \ ,  \label{eq:mixup_1} 
    \end{align}
where $\lambda \in [0,1]$ is a mixing parameter that controls the contribution of each sample. The contrastive loss function for Mixup is defined as:
\begin{align}  
  \ell_{\text{Mixup}}(\mathbf{z}_i^{(1)}, \tilde{\mathbf{z}}_i, \mathbf{z}_i^{(2)})
   = & -\lambda \log \frac{\exp 
    \left(\text{sim}\left(\tilde{\mathbf{z}}_i, \mathbf{z}_i^{(1)}\right) / {\tau}\right)}{\sum_{j=1}^B\left(\exp \left(\text{sim} \left(\tilde{\mathbf{z}}_{i}, \mathbf{z}_j^{(1)}\right) / {\tau}\right)+\exp \left(\text{sim}\left(\tilde{\mathbf{z}}_i, \mathbf{z}_j^{(2)}\right) / {\tau}\right)\right)}  \nonumber \\
      & -(1-\lambda) \log \frac{\exp \left(\text{sim}\left(\tilde{\mathbf{z}}_i, \mathbf{z}_i^{(2)}\right) / {\tau}\right)}
    {\sum_{j=1}^B\left(\exp \left(\text{sim} \left(\tilde{\mathbf{z}}_{i}, \mathbf{z}_j^{(1)}\right) / {\tau}\right)+\exp \left(\text{sim}\left(\tilde{\mathbf{z}}_i, \mathbf{z}_j^{(2)}\right) / {\tau}\right)\right)} , \\
    \mathcal{L}_{\text{SSL-Mixup}} &= \frac{1}{B} \sum_{i}^{B} \ell_{\text{Mixup}}(\mathbf{z}_i^{(1)}, \tilde{\mathbf{z}}_i, \mathbf{z}_i^{(2)}).
    \label{eq:contra_mixup}
\end{align}
 
From a mathematical view, the contrastive loss is optimized to make the similarity between original features and mixed features to retain the data mixture patterns defined in Eq.~\eqref{eq:mixup_1} as: 
\begin{equation}
    \frac{\exp 
\left(\text{sim}\left(\tilde{\mathbf{z}}_i, \mathbf{z}_i^{(1)}\right) / {\tau}\right)}{\exp \left(\text{sim}\left(\tilde{\mathbf{z}}_i, \mathbf{z}_i^{(2)}\right) / {\tau}\right)} \approx \frac{\lambda}{1-\lambda},
\label{eq:contra_2}
\end{equation}
thus predicting the mixing parameter $\lambda$ implicitly.
Namely, optimizing the contrastive loss is a pretext task where the model is trained to retain the mixture patterns of data samples in the feature space and to capture the intrinsic relationships between mixed and original data samples~\cite{ren2022simple,shen2022mix}.
    
\subsubsection{Generative-contrastive self-supervised learning}
\textit{Generative-contrastive} SSL combines generative and contrastive pretext tasks within a unified framework~\cite{liu2021self}.
This hybrid approach leverages the benefits of generative tasks, which focus on reconstructing data, and contrastive tasks, which aim to distinguish real data from synthetic data.
Among various generative-contrastive SSL methods~\cite{goodfellow2014generative,makhzani2015adversarial,wu2023self,qi2023contrast}, the generative adversarial network (GAN)~\cite{goodfellow2014generative} is one of the most prominent and widely used techniques.

As illustrated in Figure~\ref{fig:ssl_gan}, a GAN employs a generator-discriminator architecture. The generator $G$ aims to generate synthetic (`fake') data samples $\hat{\mathbf{x}}$ from random noise features $\mathbf{n}$, expressed as:
\begin{equation}
    \hat{\mathbf{x}} = G(\mathbf{n}),
\end{equation}
while the discriminator $D^{\prime}$ tries to distinguish them from real ones.
The contrastive-generative loss function is as:
\begin{equation}
    \ell_{\text{GAN}}(\mathbf{x}_i, \hat{\mathbf{x}}_i) = -\log D^{\prime}(\mathbf{x}_i)-\log \left(1-D^{\prime}(\hat{\mathbf{x}}_i)\right),
\end{equation}
\begin{equation}
    \mathcal{L}_{\text{SSL-GAN}} = \min _G \max_{D^{\prime}} \frac{1}{B} \sum_{i=1}^B \ell_{\mathrm{GAN}}\left(\mathbf{x}_i, \hat{\mathbf{x}}_i\right).
    \label{eq:gan}
\end{equation}
During pre-training, the generator $G$ is trained to minimize $\mathcal{L}_{\text{SSL-GAN}}$, improving the quality of synthetic samples; while the discriminator $D^{\prime}$ is trained to maximize $\mathcal{L}_{\text{SSL-GAN}}$, enhancing its ability to differentiate between real and fake samples.
By combining generative and contrastive tasks, GAN learns representative and discriminative features from data.
 
\subsection{Fine-tuning}
\label{sec:fine-tuning}
After pre-training, the encoder in SSL models is further transferred to the data anomaly detection (classification) tasks.
As illustrated in Stage 3 of Figure~\ref{fig:ssl_framework}, a multi-layer perception (MLP) is appended to the pre-trained encoder $E$ to form a classier, denoted as $C$.
The classifier takes a data sample $\mathbf{x}$ as input and predicts the probability $\hat{\mathbf{y}}$ over $K$ classes, which is expressed as:
\begin{equation}
    \hat{\mathbf{y}} = C(\mathbf{x}) = [\hat{y}_1, ..., \hat{y}_K],
\end{equation}
where $\hat{y}_k$ denotes the probability of class $k$.
During fine-tuning, the classifier's parameters are updated to minimize the cross-entropy on the labeled dataset, which is the commonly used loss function in classification tasks.
The cross-entropy loss function is defined as:
\begin{equation}
    \mathcal{L}_{\text{Classification}}=-\sum_{k=1}^K \mathbbm{1}_{k=y_i} \log \left(\hat{y}_{i k}\right),
    \label{eq:ce}
\end{equation}
where $\mathbbm{1}_{k=y_i}$ is the indicator function that equals 1 if predicted class $k$ is the same as true label $y_i$ and 0 otherwise; $\hat{y}_{ik}$ is the output probability that data sample $\mathbf{x}_i$ belongs to class $k$.

\label{}

\section{Experiment}
To validate the effectiveness of SSL techniques for data anomaly detection, we conduct extensive experiments using SHM data collected from real structural systems. 
The structural systems involve two in-service bridges and are introduced in Section~\ref{Experiment:1}. 
Secondly, the evaluation metrics are detailed in Section~\ref{Experiment:2}. 
Additionally, baseline SSL techniques for comparative analysis are described in Section~\ref{sec:baseline}.
The experimental results and discussion are presented in Section~\ref{Experiment:3}.

\subsection{Dataset description and preparation}
\label{Experiment:1}

\subsubsection{Case 1 - SHM data of a long-span cable-stayed bridge}
\label{sec:Case1}
\begin{figure}[!htb]
    \centering
    \includegraphics[width=0.8\linewidth]{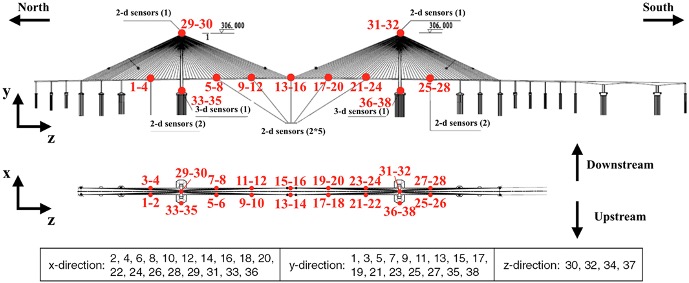}
    \caption{Sensor network of the bridge in Case 1 (image credits~\cite{tang2019convolutional}).}
    \label{fig:bridge1}
\end{figure}
Case 1 involves a long-span cable-stayed bridge in China~\cite{tang2019convolutional}.
As shown in Figure~\ref{fig:bridge1}, the bridge SHM system is equipped with 38 accelerometers, each with a sampling frequency of 20 Hz. These accelerometers measure the response acceleration of the bridge in different directions and locations.

\begin{figure*}[!htb]
    \centering
    \begin{subfigure}{.33\textwidth}
        \centering
        \includegraphics[width=\linewidth]{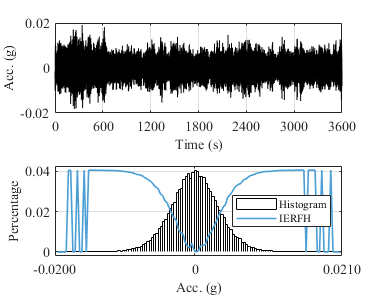} 
        \caption{Normal}
    \end{subfigure}%
    \centering
    \begin{subfigure}{.33\textwidth}
        \centering
        \includegraphics[width=\linewidth]{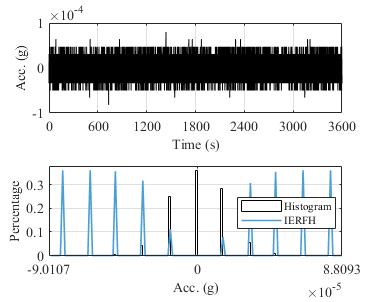} 
        \caption{Minor}
    \end{subfigure}%
    \begin{subfigure}{.33\textwidth}
        \centering
        \includegraphics[width=\linewidth]{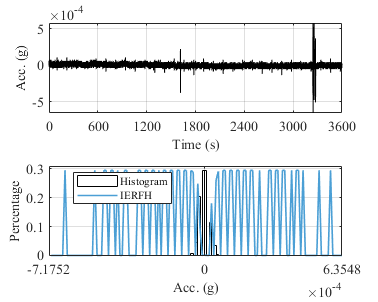} 
        \caption{Outlier}
    \end{subfigure}
    
    \begin{subfigure}{.33\textwidth}
        \centering
        \includegraphics[width=\linewidth]{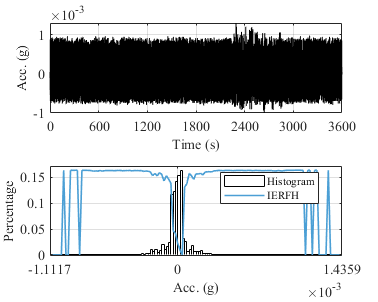} 
        \caption{Square}
    \end{subfigure}%
    \begin{subfigure}{.33\textwidth}
        \centering
        \includegraphics[width=\linewidth]{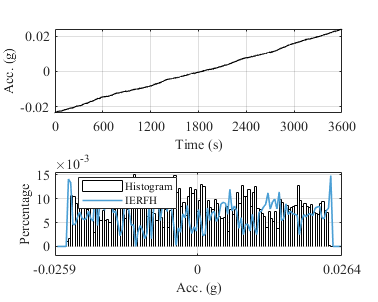} 
        \caption{Trend}
    \end{subfigure}%
    \begin{subfigure}{.33\textwidth}
        \centering
        \includegraphics[width=\linewidth]{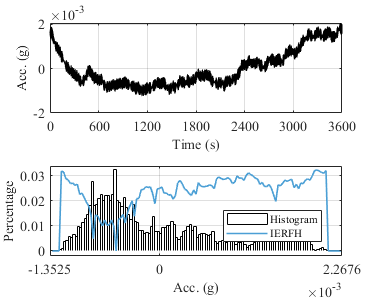} 
        \caption{Drift}
    \end{subfigure}
    \caption{Illustration of SHM data patterns in Case 1.}
    \label{fig:abnormal_type_1}
\end{figure*}

The monitoring data utilized in this work is sourced from the IPC-SHM-2020 dataset (1st International Project Competition for Structural Health Monitoring)~\cite{bao20211st}.
The dataset includes one month (January 2012) of acceleration measurements.
The raw continuous data is split into 1-hour segments using non-overlapping sliding windows, resulting in a total of 28,272 samples (31 days $\times$ 24 h $\times$ 38 channels).
Each 1-hour segment contains 72,000 data points (3600 s $\times$ 20 Hz), which are then transformed into a 512-dimensional IERFH feature using the data reduction method described in Section~\ref{sec:data_reduction}.
The data patterns are defined by domain experts. A detailed description of each data pattern is provided in Table~\ref{tab:da_description}, and Figure~\ref{fig:abnormal_type_1} illustrates examples of these data patterns. 

\subsubsection{Case 2 - SHM data of a long-span arch bridge}
\begin{figure}[!htb]
    \centering
    \includegraphics[width=0.7\linewidth]{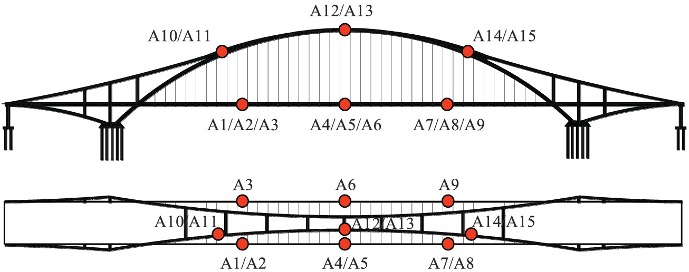}
    \caption{Sensor network of the bridge in Case 2 (image credits~\cite{jian2021faulty}).}
    \label{fig:bridge2}
\end{figure}

In Case 2, we validate the SSl-based framework using data from a long-span arch bridge in China~\cite{jian2021faulty}, a half-through tied-arch bridge that has been in service since 2011.
As illustrated in Figure~\ref{fig:bridge2}, the SHM system of this bridge consists of 15 single-axis accelerometers, each with a sampling rate of 50 Hz and a measurement range from -50 and 50 mg (1 mg = 0.01 m/$s^2$).

\begin{figure*}[!htb]
    \centering
    \begin{subfigure}{.33\textwidth}
        \centering
        \includegraphics[width=\linewidth]{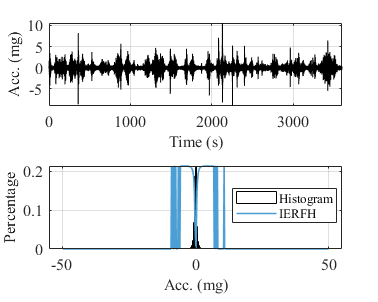} 
        \caption{Normal}
    \end{subfigure}%
    \begin{subfigure}{.33\textwidth}
        \centering
        \includegraphics[width=\linewidth]{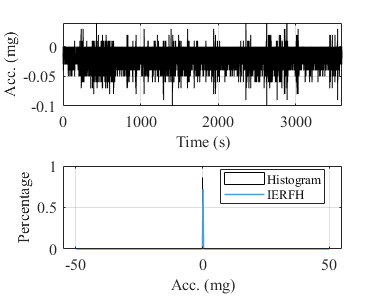} 
        \caption{Minor}
    \end{subfigure}%
    \begin{subfigure}{.33\textwidth}
        \centering
        \includegraphics[width=\linewidth]{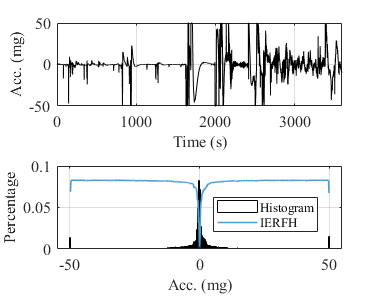} 
        \caption{Outlier}
    \end{subfigure}
    
    \begin{subfigure}{.33\textwidth}
        \centering
        \includegraphics[width=\linewidth]{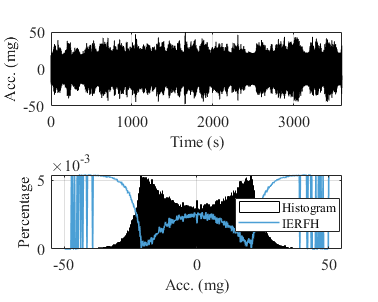} 
        \caption{Noise-corrupted}
    \end{subfigure}%
    \begin{subfigure}{.33\textwidth}
        \centering
        \includegraphics[width=\linewidth]{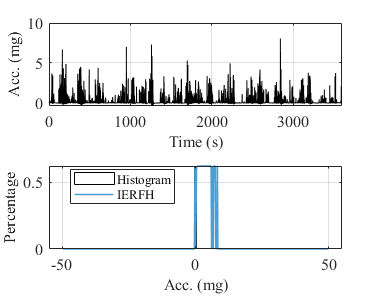} 
        \caption{Biased}
    \end{subfigure}
    \caption{Illustration of SHM data patterns in Case 2.}
    \label{fig:abnormal_type_2}
\end{figure*}

The utilized SHM data is collected over three months (November 2018, December 2018, and June 2019).
Following the same windowing procedure in Case 1, the raw data is divided into 34,920 samples (97 days $\times$ 24 h $\times$ 15 channels).
Each 1-hour segment contains 180,000 data points (3600 s $\times$ 50 Hz), which are subsequently transformed in 512-dimensional IERFH feature using the data reduction technique described in Section~\ref{sec:data_reduction}.
Table~\ref{tab:da_description} lists the data patterns defined via expert knowledge. 
Examples of these data patterns are illustrated in Figure~\ref{fig:abnormal_type_2}.
\begin{table*}[!htb] 
    \centering
    \caption{Description of data patterns of the SHM data.}
    \begin{tabular}{llcc}
        \hline
        Data patterns  & Description & Case 1 & Case 2\\
        \hline
        Normal & Data without abnormal pattern & $\checkmark$ & $\checkmark$ \\
        Missing & Data with partially or whole missing measurements & $\checkmark$ & $\checkmark$\\
        Minor & Data's amplitude is very small in the time domain, compared with normal data & $\checkmark$ & $\checkmark$\\
        Outlier & Data include outliers, like spikes or abnormal fluctuations, etc. & $\checkmark$ & $\checkmark$ \\
        Square & Data exhibit a square wave shape   & $\checkmark$ & -\\
        Trend & \makecell[l]{Data consist of trend item and has the peak value in frequency domain\\(mean centering applied)}  & $\checkmark$ & -\\
        Drift &Data are non-stationary, with random drifts (mean centering applied)  & $\checkmark$ & -\\
        Biased & Data's time history is asymmetric with respect to the time axis  & - & $\checkmark$ \\
        Noise & Data are corrupted by irregular noise  & - & $\checkmark$\\
        \hline
    \end{tabular}
    \label{tab:da_description}
\end{table*}

\subsubsection{Dataset preparation}
In each case, the acquired SHM data is split into three subsets: labeled, validation, and test datasets, with proportion ratios of 20\%, 30\%, and 50\%, respectively.
Model performance is evaluated on the test dataset after training, while the optimal model is determined using the validation dataset during training.

Our work aims to develop a data anomaly detection method with limited labeled data.
To this end, we create labeled \textit{low-shot} datasets from the labeled dataset.
These low-shot datasets contain a small number of labeled samples for each abnormal pattern, reflecting the situation where label information is scarce.
Since abnormal data in SHM is often imbalanced, deep learning models are typically trained or fine-tuned using both balanced and unbalanced labeled datasets, as suggested in previous studies~\cite{pan2023transfer,tang2019convolutional}.
In this study, we design six labeled low-shot datasets, considering both balanced and unbalanced conditions. The details of these low-shot labeled datasets are summarized in Table~\ref{tab:Case1} and Table~\ref{tab:Case2}. 

It is noted that the `missing' abnormal pattern is excluded from anomaly detection in this study. This pattern can be readily identified by counting the number of data points in software~\cite{jian2021faulty}. 
Therefore, we don't consider the `missing' pattern in training and testing to save computational resources. 
After removing the `missing' pattern, the original dataset for Case 1 consists of 25,330 data samples, while Case 2 includes 28,292 data samples.
For SSL pre-training, neural networks are trained on the full original dataset without access to label information. After pre-training, the low-shot labeled datasets are used to fine-tune the pre-trained SSL model.
\begin{table}[!htb]
    \centering
    \caption{Configuration of datasets in Case 1.}
    \begin{tabular}{lcccccccc}
        \hline
        Dataset setting  & Normal  & Minor & Outlier & Square & Trend & Drift & Total number  \\
        \hline
        Original, $\mathcal{D}^1$  & 13575  & 1775& 527&2996&5778&679 & 25330  \\
        \hline
        Test, $\mathcal{D}_{\text{test}}^1$ & 6787 & 887 & 263 & 1498 & 2889 & 339 & 12743 \\
        Validation, $\mathcal{D}_{\text{val}}^1$ & 4073 & 533 & 158 & 898 & 1733 & 204 & 7506  \\
        Label, $\mathcal{D}_{\text{label}}^1$ & 2715 & 355 & 106 & 600 & 1160 & 136 & 5071 \\
        \hline
        \textit{Balanced labeled dataset} \\
         Low-shot 1, $\mathcal{D}_{1}^1$ & 10  & 10 & 10 & 10 & 10 & 10 & 60  \\
         Low-shot 2, $\mathcal{D}_{2}^1$ & 30 & 30 & 30 &30 &30 &30 & 180  \\
         Low-shot 3, $\mathcal{D}_{3}^1$ & 50 & 50 & 50 &50 &50 &50 & 300  \\
         \hline
        \textit{Unbalanced labeled dataset} \\
        Low-shot 4, $\mathcal{D}_{1}^1$ & 50 & 30 & 30 & 30 & 50 & 30 & 220  \\
        Low-shot 5, $\mathcal{D}_{2}^1$ & 100 & 50 & 50 & 50 & 80 & 50 & 380  \\
        Low-shot 6, $\mathcal{D}_{3}^1$ & 200 & 50 & 50 & 50 & 150 & 50 & 550  \\
        \hline
    \end{tabular}
    \label{tab:Case1}
\end{table}
\begin{table}[!htb]
    \centering
    \caption{Configuration of datasets in Case 2.}
    \begin{tabular}{lccccccc}
        \hline
        Dataset  & Normal & Minor & Biased & Outlier & Noise & Total number  \\
        \hline
        Original dataset, $\mathcal{D}^2$  & 19454  & 6802 & 1169 & 289 & 578 & 28292 \\
        \hline
        Test, $\mathcal{D}_{\text{test}}^2$ & 9727 & 3401 & 584 & 144 & 209 & 14065 \\
        Validation, $\mathcal{D}_{\text{val}}^2$ & 5836 & 2040 & 351 & 87 & 173 & 8487  \\
        Label, $\mathcal{D}_{\text{label}}^2$ & 3891 & 1361 & 234 & 58 & 116 & 5660  \\
        \hline
        \textit{Balanced labeled dataset} \\
        Low-shot 1, $\mathcal{D}_{1}^2$ &10 &10&10&10 &10 & 50 \\
        Low-shot 2, $\mathcal{D}_{3}^2$ &30 &30&30&30&30 & 150  \\
        Low-shot 3, $\mathcal{D}_{3}^2$ &50 &50&50&50&50 & 250 \\
        \hline
        \textit{Unbalanced labeled dataset} \\
        Low-shot 4, $\mathcal{D}_{4}^2$ & 30 & 20 & 20 & 10 & 10 & 90 \\
        Low-shot 5, $\mathcal{D}_{5}^2$ & 50 & 50 & 30 & 30 & 30 & 190 \\
        Low-shot 6, $\mathcal{D}_{6}^2$ & 100 & 80 & 50 & 50 & 50 & 330  \\
        \hline
    \end{tabular}
    \label{tab:Case2}
\end{table}

\subsection{Evaluation metrics}
\label{Experiment:2}
In this work, SHM data anomaly detection is formulated as a classification task.
For such tasks, the confusion matrix is a widely used tool that records the relationship between the true and predicted classes of data samples~\cite{grandini2020metrics}.
Table~\ref{table:cm} presents an example of a confusion matrix for a binary classification task. In this matrix:
TP (true positive) are data samples correctly predicted to be the target class;
FP (false positive) are data samples incorrectly predicted as the target classes;
TN (true negative) are data samples correctly predicted to be other classes;
FN (false negative) are data samples misclassified as other classes. 
\begin{table*}[!htb]
    \caption{Illustration of confusion matrix.}
\centering
    \begin{tabular}{cccccc}
        \hline
        & \multirow{2}{*}{Confusion matrix} & &  \multicolumn{2}{c}{Predicted result} \\
         \cline{4-5} 
        & & & Positive & Negative \\ 
        \hline
        & \multirow{2}{*}{Real label} & Positive & True positive (TP) & False negative (FN)\\
        & & Negative & False positive (FP) & True negative (TN)\\
        \hline
    \end{tabular}
    \label{table:cm}
\end{table*}
Based on the confusion matrix, typical evaluation metrics for classification tasks are defined as follows:
\begin{align}
    \label{eq:class_metric_precision}
    & \mathrm{Precision = \frac{TP}{TP+FP}} \ , \\
    \label{eq:class_metric_recall}
    & \mathrm{Recall = \frac{TP}{TP+FN}} \ , \\
    \label{eq:class_metric_acc}
    & \mathrm{Accuracy = \frac{TP+TN}{TP+TN+FP+FN}} \ , \\
    \label{eq:class_metric_f1}
    & \mathrm{F_1 \:Score = 2 \cdot \frac{Precision \cdot Recall}{Precision + Recall}} \ . 
\end{align}
Among these metrics, we use the $F_1$ score as the overall performance metric, as it balances precision and recall, and it is also suggested in previous studies~\cite{pan2023transfer,tang2019convolutional}. 
Additionally, accuracy is reported to show the classification performance.
Given that SHM data distributions are often imbalanced, with varying amounts of data for each abnormal pattern, we also present the precision and recall for each class, enabling us to evaluate the model performance at class level.

\subsection{Baseline methods}
\label{sec:baseline}
To explore the effective SSL method for SHM data anomaly detection, we conduct a comparative analysis with four mainstream SSL methods alongside purely supervised training.
This work focuses on detecting anomalies with very limited labeled data. 
To ensure a fair comparison, all methods are provided with the same amount of labeled data. The baseline methods are as follows:

\begin{itemize}
    \item Supervised training (SUP):
    Supervised classification is a standard approach commonly used in SHM data anomaly detection.
    In this method, a classifier $C$ is trained by minimizing cross-entropy in Eq.~\eqref{eq:ce} on the labeled low-shot dataset. The classifier architecture is identical to the one in the SSL framework.
    This method provides a baseline to assess the impact of SSL pre-training compared to purely supervised learning.
    \item Autoencoder (AE):
    AE is a \textit{generative} SSL method that reconstructs input data using an encoder-decoder architecture.
    The encoder $E$ is the default CNN-based encoder (detailed below), while the decoder $D$ mirrors the encoder structure. AE is trained via the reconstruction loss in Eq.~\eqref{eq:ae}.
    Data augmentation and additional hyperparameters are not required for this method.
    \item Simple contrastive learning (SimCLR): SimCLR is a \textit{contrastive} SSL method based on an encoder-projector architecture. Following the suggested settings in~\cite{chen2020simple}, we apply crop operation and add Gaussian noise as augmentation transformations. The temperature parameter $\tau$ is set to 0.5. The encoder E and the projector $P$ are optimized via the loss function in Eq.~\eqref{eq:ssl_simclr}. 
    \item Mixup contrastive learning (Mixup): Mixup is another \textit{contrastive} SSL method. The employed architecture and loss function are based on~\cite{wickstrom2022mixing}. Mixup data augmentation generates synthetic samples by combining pairs of data samples using a mixing parameter $\lambda$, sampled from a beta distribution Beta$(0.2,0.2)$. The temperature parameter $\tau$ is set to 0.1.
    The encoder E and the projector $P$ are optimized via the loss function in Eq.~\eqref{eq:contra_mixup}. 
    \item Generative adversarial network (GAN): GAN is a \textit{generative-contrastive} SSL method that learns latent features through adversarial training. A generator-discriminator architecture is optimized using the loss function in Eq.~\eqref{eq:gan}.
    Data augmentation and additional hyperparameters are not required for this method.
 \end{itemize}
 
The encoder $E$ is a CNN with 5 layers, each containing a 1D convolution, batch normalization, and ReLU activation. The channel numbers for the layers are [16, 32, 64, 128, 256], with kernel sizes of [5, 3, 3, 3, 3] and strides of [5, 3, 3, 3, 3], respectively.
The MLP appended to the pre-trained encoder consists of three layers with input, hidden, and output dimensions of $[256, 256, K]$, with ReLU activation in the hidden layer and $K$ representing the number of data patterns.
The decoder $D$ includes 5 layers, each containing a 1D deconvolution, batch normalization, and ReLU activation.
The channel numbers for the layers are [128, 64, 32, 16, 1], with kernel sizes of [3, 3, 3, 3, 5] and strides of [3, 3, 3, 3, 5], respectively.
The projector $P$ is a single linear layer that maps 256-dimensional input to a 128-dimensional output.
The generator $G$ has the same architecture as the decoder.
The discriminator $D^{\prime}$ consists of a CNN-based encoder and a linear layer, which amps 256-dimensional input to a single output.

Deep learning models are developed using the PyTorch~\cite{paszke2019pytorch} deep learning library, and experiments are conducted on an NVIDIA Geforce RTX 3080 GPU. 
The Adam~\cite{kingma2014adam} optimizer is used to optimize model parameters during SSL pre-training, supervised training, and fine-tuning stages.
The batch size is 64, and the learning rate is 0.001.
The SSL pre-training epochs are 200, and Supervised learning (SUP) is also trained for 200 epochs. During fine-tuning, the pre-trained models are further fine-tuned for 50 epochs.
For each baseline method, the fine-tuning and evaluation are implemented 5 times, and mean values and standard deviation are provided in experimental results.

\subsection{Experimental results}
\label{Experiment:3}
\subsubsection{Overall performance}
Table~\ref{table:general_1} presents the overall performance ($F_1$ scores) of the baseline methods in Case 1.
Among all methods, AE achieves an $F_1$ score improvement of approximately $3\% - 8\%$ compared to the purely supervised training (SUP) approach, given the same amount of labeled data.
Notably, AE method achieves the highest $F_1$ score of $80.52\%$ on the test dataset $\mathcal{D}_{\text{test}}^1$ using the low-shot labeled dataset $\mathcal{D}_6^1$.
SimCLR follows closely, achieving the second-best $F_1$ score of $79.32\%$ under the same conditions.
However, it is noted that GAN pre-training achieves an $F_1$ score of $74.08\%$ given labeled dataset $\mathcal{D}_6^1$, underperforming compared to SUP method. 
We term such cases of decreased performance as \textit{negative} pre-training.

Table~\ref{table:general_2} shows the $F_1$ scores of the baseline methods in Case 2.
AE pre-training method shows consistent improvement, achieving an increase of approximately $3\% - 9\%$ in $F_1$ scores compared to SUP method.
Given the low-shot labeled dataset $\mathcal{D}_6^2$, AE attains the best $F_1$ score of $89.62\%$ on the test dataset $\mathcal{D}_{\text{test}}^2$.
However, more cases of negative pre-training are observed in Case 2.
For example, SimCLR underperforms compared to SUP given labeled datasets $\mathcal{D}_2^2$ and $\mathcal{D}_5^2$, and Mixup exhibits similar issues given labeled dataset $\mathcal{D}_2^2$ and $\mathcal{D}_4^2$.
\begin{table}[!htb] 
    \caption{$F_1$ scores (\%) of different methods on Case 1, fine-tuned on various low-shot datasets.}
\centering
    \begin{tabular}{lcccccc}
        \hline
        \multirow{2}{*}{Method}  & \multicolumn{6}{c}{Low-shot labeled dataset}  \\
        \cline{2-7} 
        & $\mathcal{D}_{1}^1$ &  $\mathcal{D}_{2}^1$ &  $\mathcal{D}_{3}^1$ & $\mathcal{D}_{4}^1$ & $\mathcal{D}_{5}^1$ & $\mathcal{D}_{6}^1$ \\
        \hline 
        SUP & $66.87 \pm 1.39$ & $68.17 \pm 0.92$ & $73.50 \pm 2.21$  & $72.44 \pm 0.49$ & $75.53 \pm 0.69$ & $77.16 \pm 0.73$ \\
        \hline
        AE & $74.17 \pm 0.93$ & $74.25 \pm 0.33$ & $78.44 \pm 2.27$  & $76.39 \pm 0.54$ & $77.91 \pm 0.86$ & $\mathbf{80.52 \pm 0.29}$\\
        SimCLR & $73.45 \pm 1.71$ & $75.32 \pm 0.51$ & $77.73 \pm 0.48$ & $77.41 \pm 0.51$ & $77.73 \pm 1.10$ & $79.32 \pm 1.29$ \\
        Mixup & $71.62 \pm 1.27$ & $71.09 \pm 0.97$ & $75.89 \pm 0.48$ & $74.45 \pm 0.84$ & $76.57 \pm 0.47$ & $77.90 \pm 0.62$\\ 
        GAN & $63.44 \pm 1.82$ & $69.34 \pm 0.66$ & $74.14 \pm 0.79$ & $69.46 \pm 2.58$ & $72.97 \pm 0.68$ & $74.08 \pm 3.70$\\
         \hline
    \end{tabular}
    \label{table:general_1}
\end{table}
\begin{table}[!htb] 
    \caption{$F_1$ scores (\%) of different methods on Case 2, fine-tuned on various low-shot datasets.}
\centering
    \begin{tabular}{lcccccc}
        \hline
        \multirow{2}{*}{Method}  & \multicolumn{6}{c}{Low-shot labeled dataset}  \\
        \cline{2-7} 
        & $\mathcal{D}_{1}^2$ &  $\mathcal{D}_{2}^2$ &  $\mathcal{D}_{3}^2$ & $\mathcal{D}_{4}^2$ & $\mathcal{D}_{5}^2$ & $\mathcal{D}_{6}^2$ \\
        \hline 
        SUP & $68.31 \pm 3.16$ & $ 77.58 \pm 2.08 $ & $ 78.96 \pm 1.58 $ & $ 81.61 \pm 1.83 $ & $ 80.90 \pm 2.27 $ & $ 85.32 \pm 1.09 $ \\
        \hline
        AE  & $75.38 \pm 3.20$ & $ 82.92 \pm 1.27 $ & $ 87.93 \pm 1.81 $ & $ 84.70 \pm 1.41$ & $ 86.71 \pm 0.85 $ & $\mathbf{89.62 \pm 0.78} $\\
        SimCLR & $ 70.87 \pm 3.35 $ & $ 75.14 \pm 1.54 $ & $ 82.18 \pm 0.60 $ & $ 82.25 \pm 0.68 $ & $ 79.38 \pm 3.02 $ & $ 86.83 \pm 2.59 $\\
        Mixup & $73.86 \pm 3.03$ & $74.14 \pm 2.20$  & $ 79.96 \pm 1.86$ & $ 76.45 \pm 3.34$ & $ 82.16 \pm 0.86$& $ 86.42 \pm 0.72$  \\
        GAN & $72.69 \pm 4.39$ & $77.25 \pm 1.46$  & $ 83.05 \pm 1.10$ & $ 79.88 \pm 2.63$ & $ 80.33 \pm 0.23$& $ 88.89 \pm 0.98$\\
        \hline
    \end{tabular}
    \label{table:general_2}
\end{table}

\begin{figure}[!htb]
    \begin{subfigure}{\linewidth}
        \centering
        \includegraphics[width=0.7\textwidth]{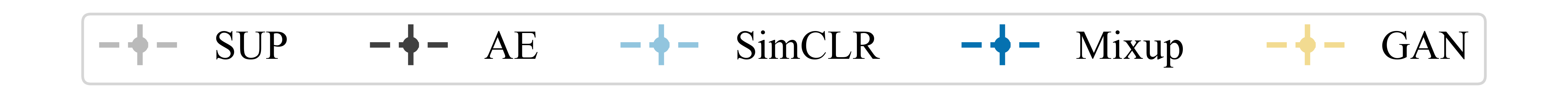}
    \end{subfigure}\\
    \begin{subfigure}{.5\linewidth}
        \centering
       \includegraphics[width=\linewidth]{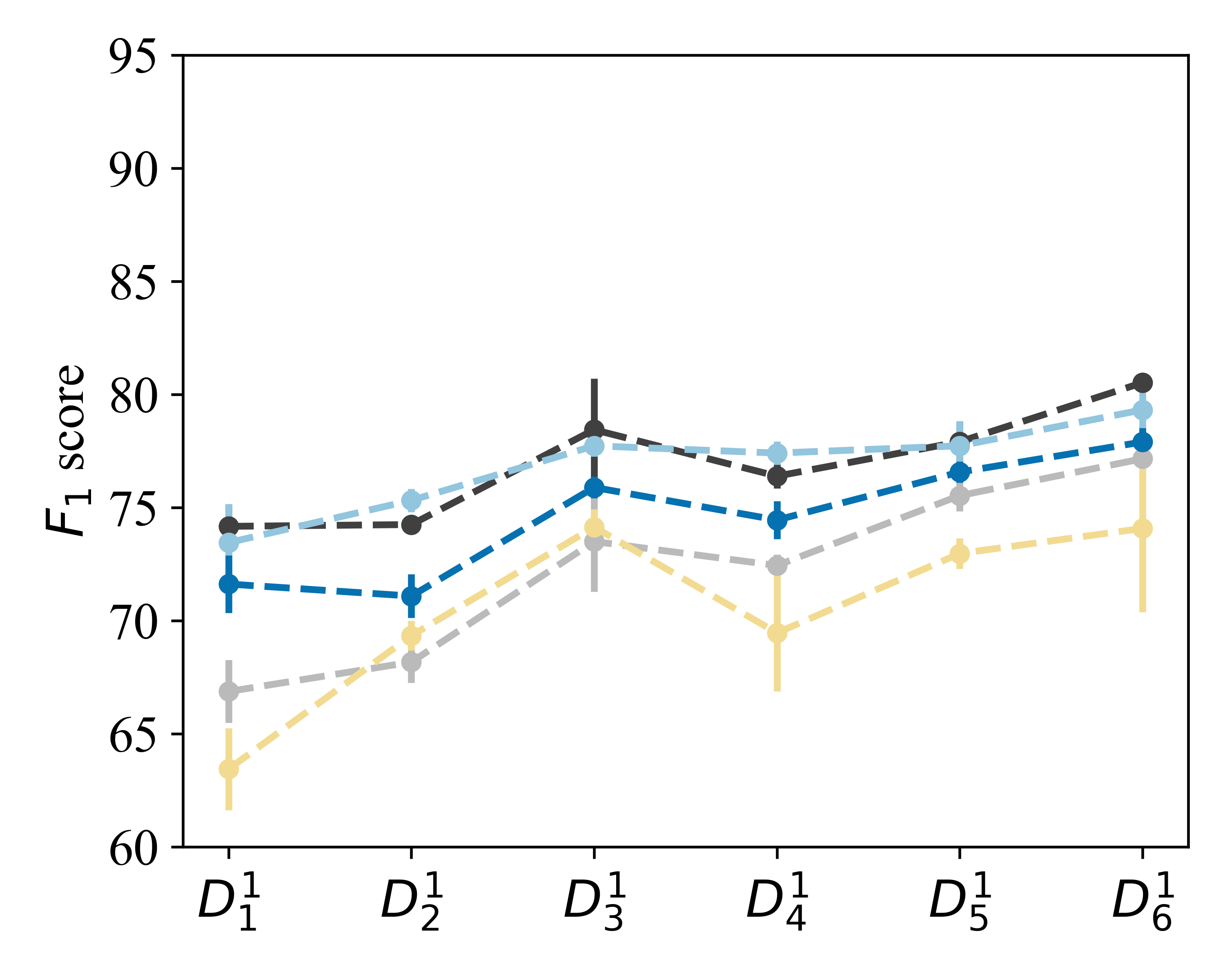}
       \caption{Results of Case 1.}
    \end{subfigure}%
    \begin{subfigure}{.5\linewidth}
        \centering
        \includegraphics[width=\linewidth]{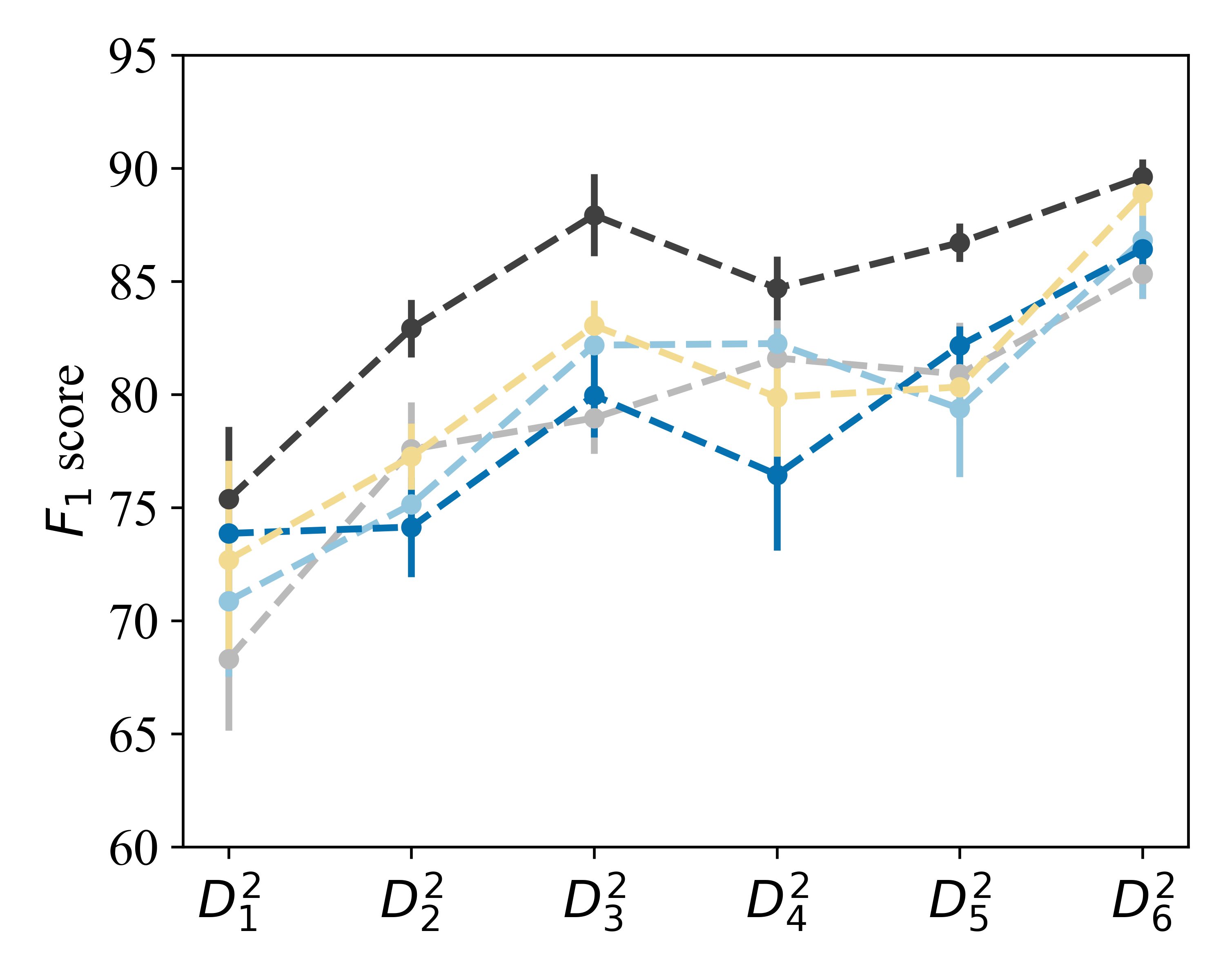}
        \caption{Results of Case 2.}
     \end{subfigure}
    \caption{Bar plot $F_1$ scores (\%) of different methods, utilizing labeled data from various low-shot datasets.}
    \label{fig:B1_bar}
\end{figure}

Figure~\ref{fig:B1_bar} provides a bar plot comparison of $F_1$ scores across different baseline methods.
In Case 1, nearly all SSL methods demonstrate improved performance compared to SUP method.
Among various SSL methods, AE and SimCLR achieve the best or the second-best $F_1$ scores across low-shot datasets. 
In Case 2, AE pre-training consistently outperforms SUP methods across all scenarios.
However, SimCLR, Mixup, and GAN pre-training occasionally result in decreased $F_1$ scores, indicating that these pre-training methods can have a negative effect under certain conditions.

\begin{figure*}[!htb]
    \centering
    \begin{subfigure}{.5\textwidth}
        \centering
        \includegraphics[width=\linewidth]{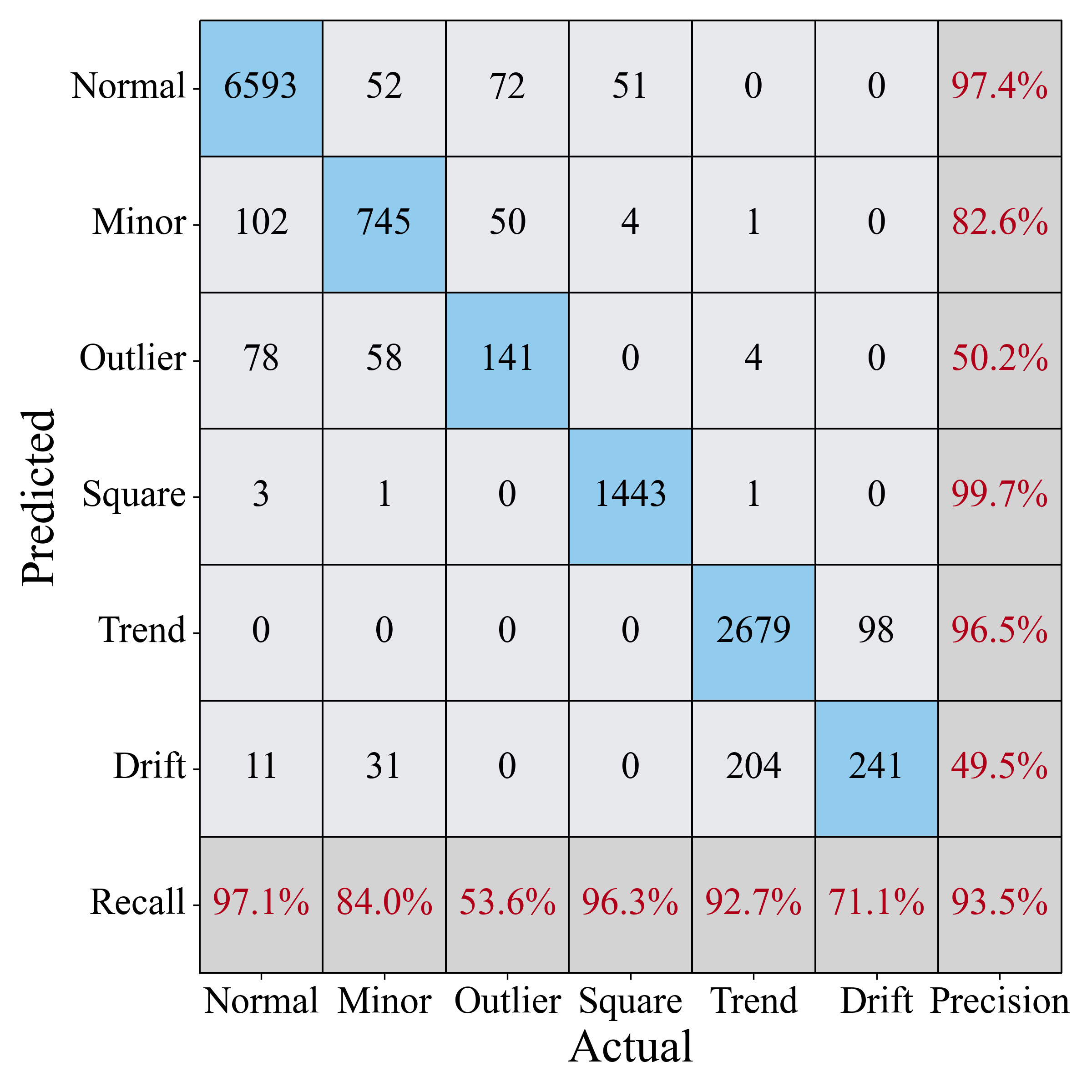} 
        \caption{Case 1: AE, fine-tuned on $\mathcal{D}_6^1$}
        \label{}
    \end{subfigure}%
    \begin{subfigure}{.5\textwidth}
        \centering
        \includegraphics[width=\linewidth]{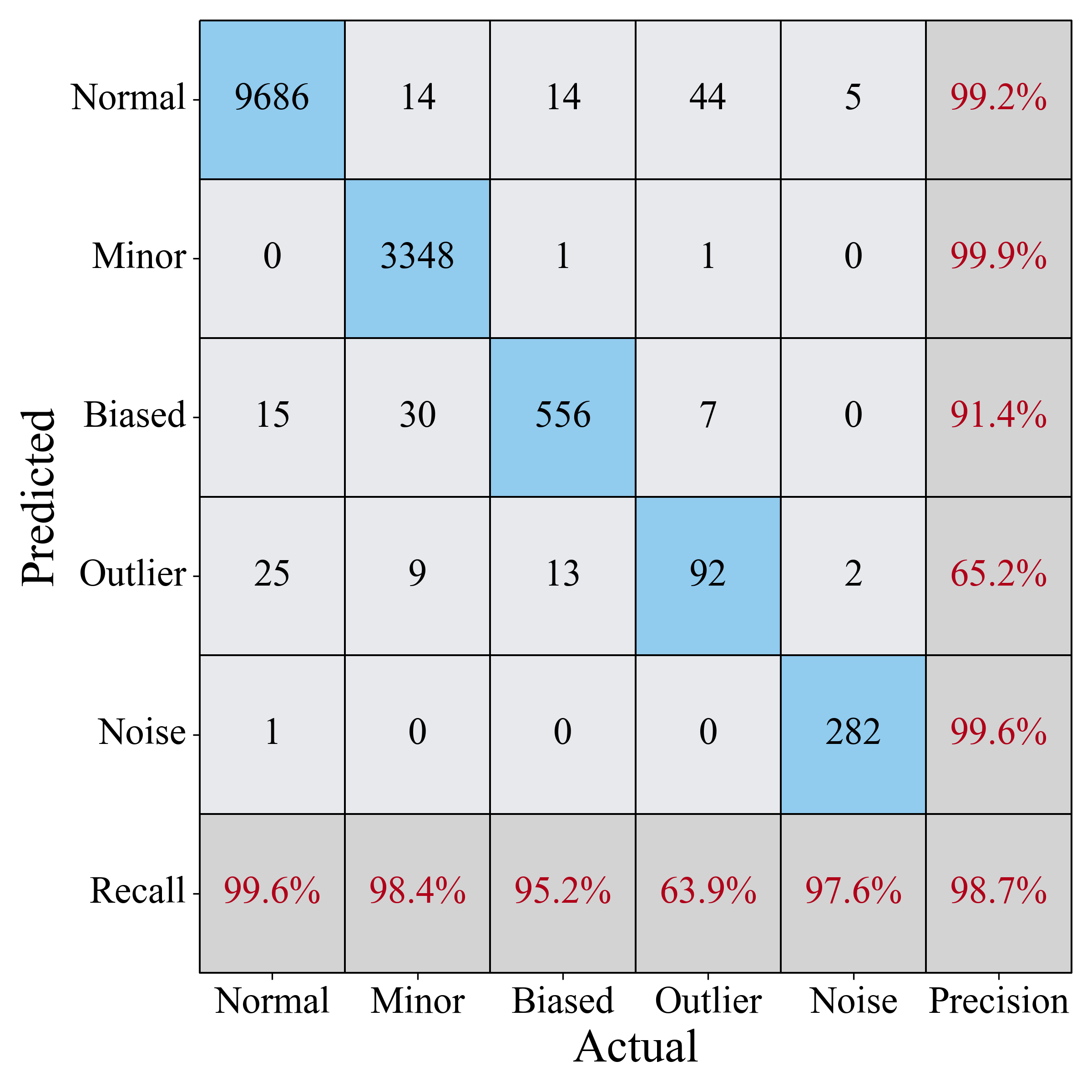} 
        \caption{Case 2: AE, fine-tuned on $\mathcal{D}_6^2$}
    \end{subfigure}
    \caption{Confusion matrices of AE method on test datasets (numbers in the bottom right corner represent the accuracy).}
    \label{fig:cfm}
\end{figure*}

We also report the overall accuracy of the best-performing SSL method.
Figure~\ref{fig:cfm} shows the confusion matrices of AE method on the test datasets from two validation cases.
AE achieves an overall accuracy of $93.5\%$ in Case 1 and $98.7\%$ in Case 2. These results highlight the effectiveness of AE pre-training in anomaly detection tasks.

In summary, given limited labeled data, SSL pre-training, particularly AE method, demonstrates
clear effectiveness and consistent performance improvements in SHM data anomaly detection.
Despite some instances of negative pre-training with other SSL methods, AE emerges as the most robust approach, achieving the best overall $F_1$ scores and accuracy across both validation cases.

\subsubsection{Class level performance}
For anomaly detection methods, accurately predicting `normal' data is crucial, as these predictions are treated as clean and reliable data for subsequent SHM analysis.
As shown in Figure~\ref{fig:cfm}, AE achieves high precision on `normal' data, with $97.4 \%$ in Case 1 and $99.2 \%$ in Case 2.
For abnormal data, the precision and recall of the majority of abnormal patterns, such as `square' and `trend' (in Case 1), `minor', `biased', and `noise' (in Case 2), are generally above $90\%$. However, for minority abnormal patterns like `outlier' (in Case 1\&2) and `drift' (in Case 1), the precision and recall fall below $90\%$, indicating a huge gap in performance for these less frequent data patterns.

Overall, AE pre-training method exhibits very good performance in identifying `normal' data and detecting the majority of abnormal patterns, making it highly valuable for data cleansing in SHM practices.
However, all current methods show limitations in accurately detecting minority patterns.
Developing effective methods for these minority abnormal data is still an open and challenging problem for further research.

\subsubsection{Pre-training and fine-tuning stages}
\begin{figure*}[!htb]
    \begin{subfigure}{.5\textwidth}
        \centering
        \includegraphics[width=\linewidth]{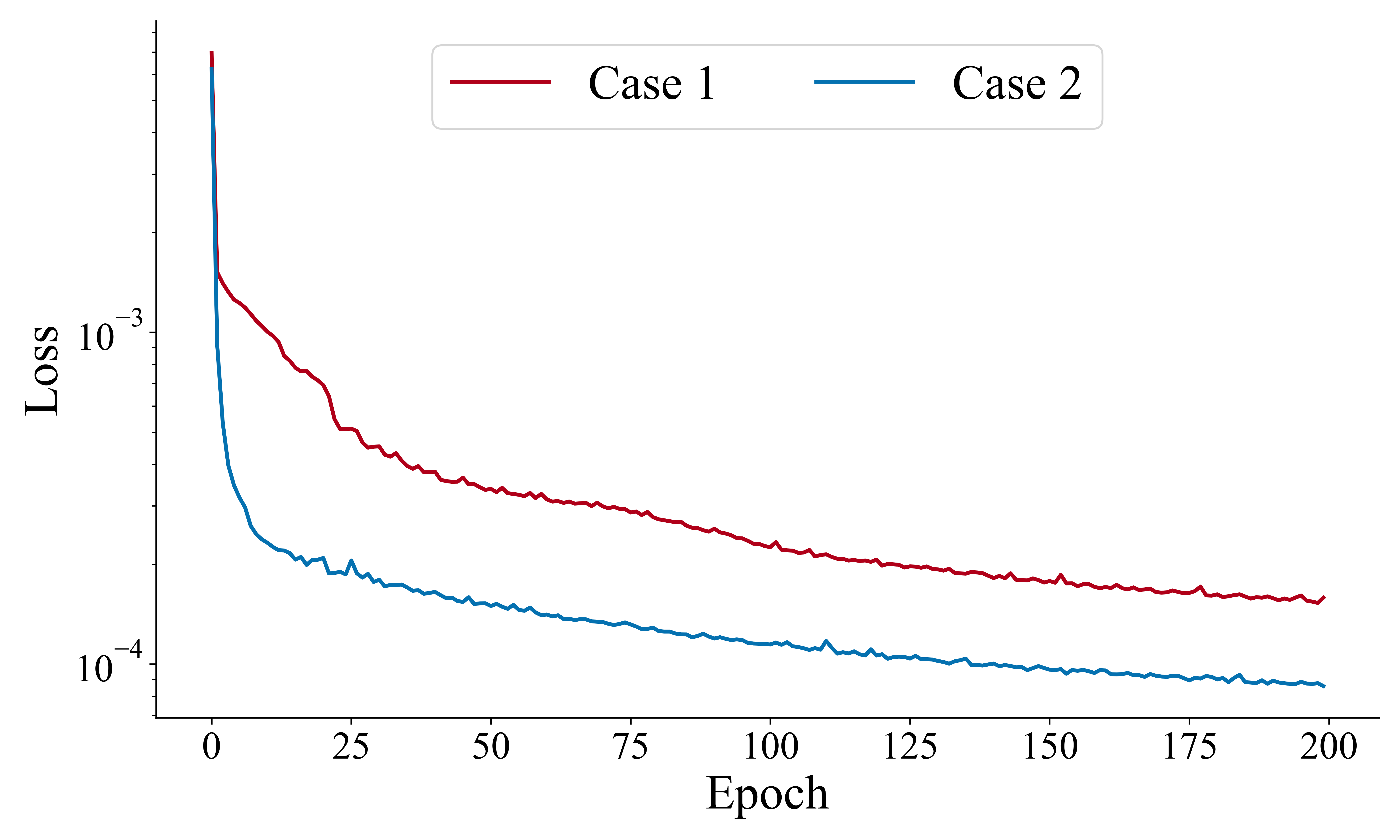} 
        \caption{AE}
    \end{subfigure}%
    \begin{subfigure}{.5\textwidth}
        \centering
        \includegraphics[width=\linewidth]{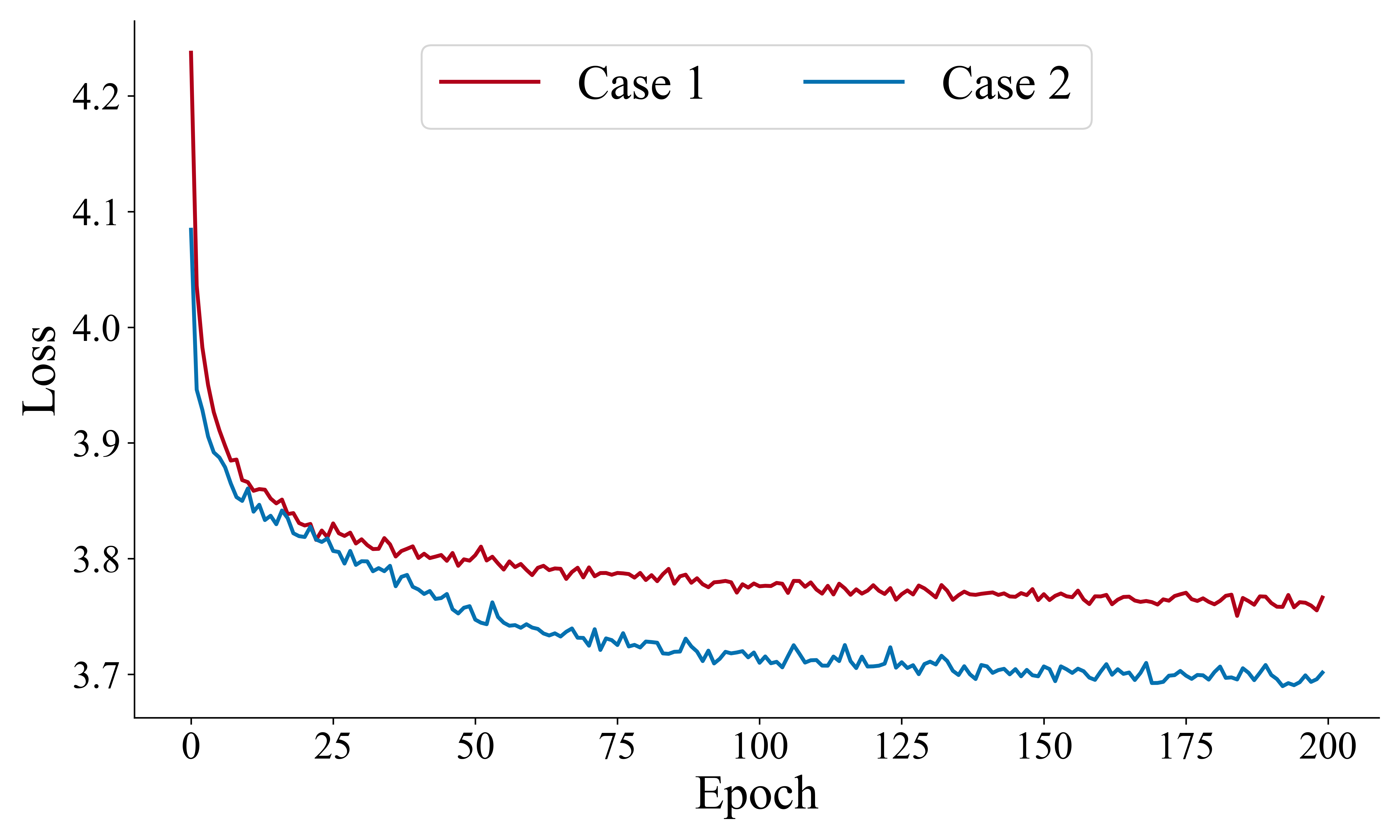} 
        \caption{SimCLR}

    \end{subfigure}
    
    \begin{subfigure}{.5\textwidth}
        \centering
        \includegraphics[width=\linewidth]{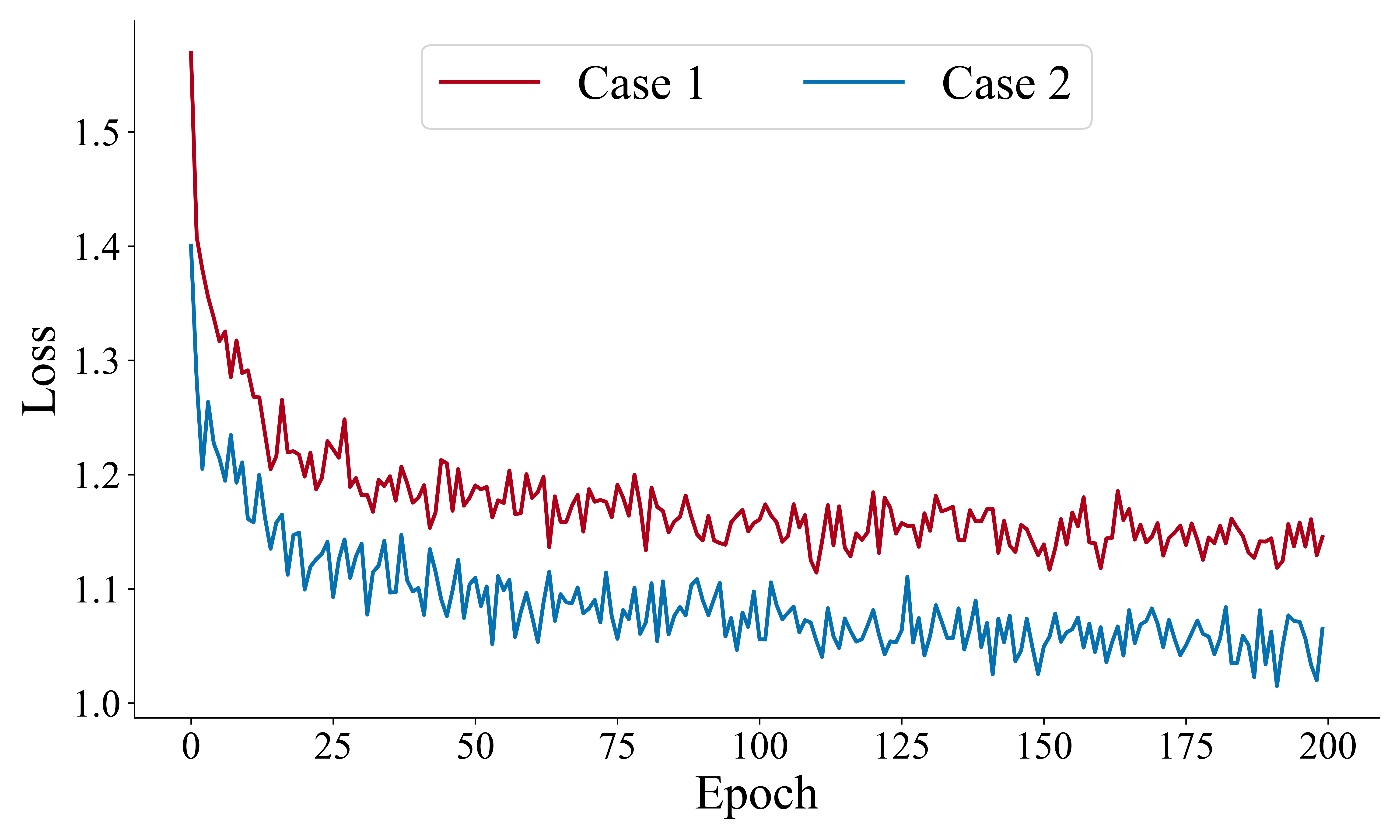} 
        \caption{Mixup}
    \end{subfigure}%
    \begin{subfigure}{.5\textwidth}
        \centering
        \includegraphics[width=\linewidth]{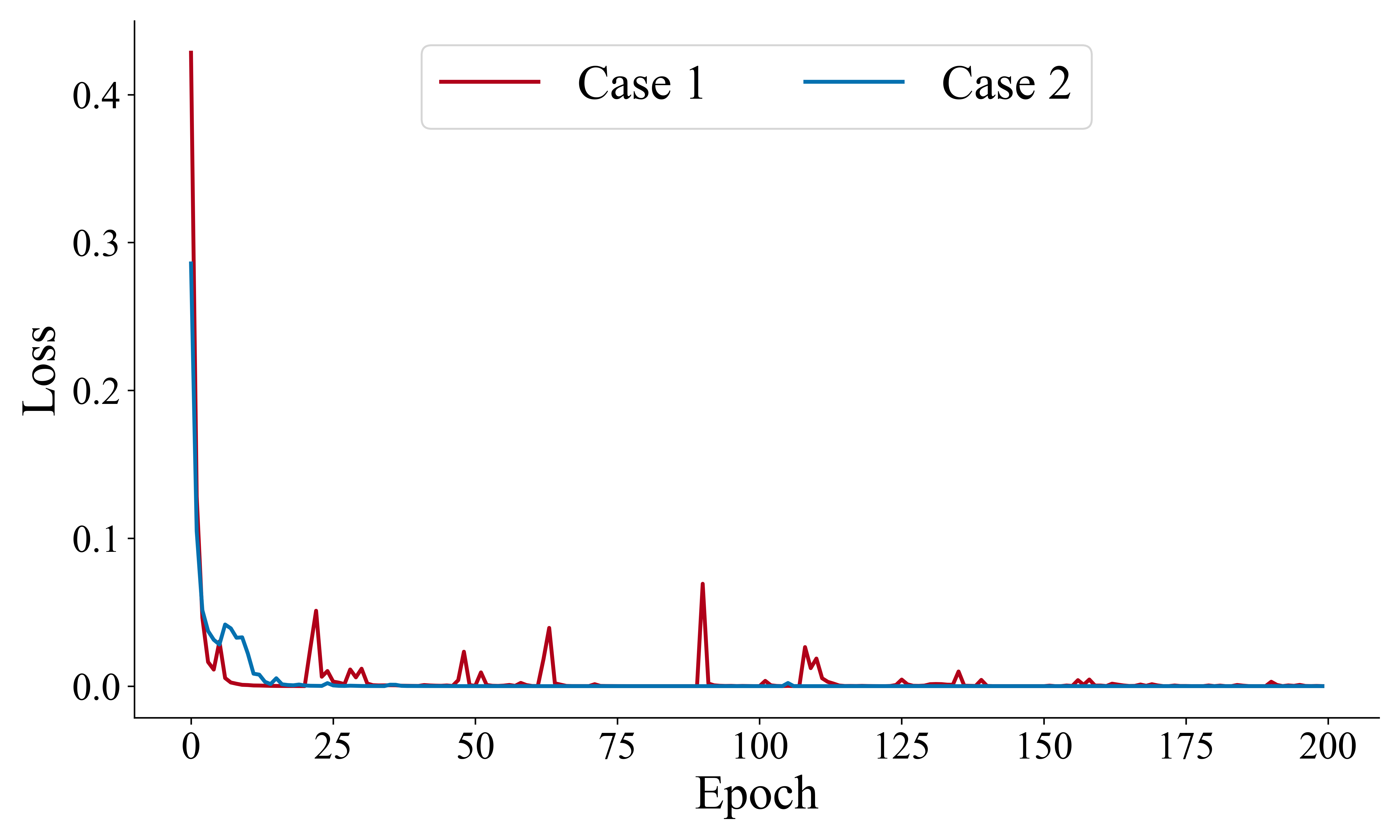} 
        \caption{GAN}
    \end{subfigure}
    \caption{Loss functions during the self-supervised pre-training stage.}
    \label{fig:loss}
\end{figure*}

Figure~\ref{fig:loss} presents the loss functions of various SSL methods during the pre-training stage.
The results show that the defined SSL loss functions are effectively minimized during pre-training, indicating that the neural networks successfully learn data representations aligned with the properties defined by pretext tasks (\textit{generative}, \textit{contrastive} or \textit{generative-contrastive}).
It is noted that since the SSL loss functions correspond to different pretext tasks, the scales in each subfigure are different.

\begin{figure*}[!htb]
    \begin{subfigure}{.5\textwidth}
        \centering
        \includegraphics[width=\linewidth]{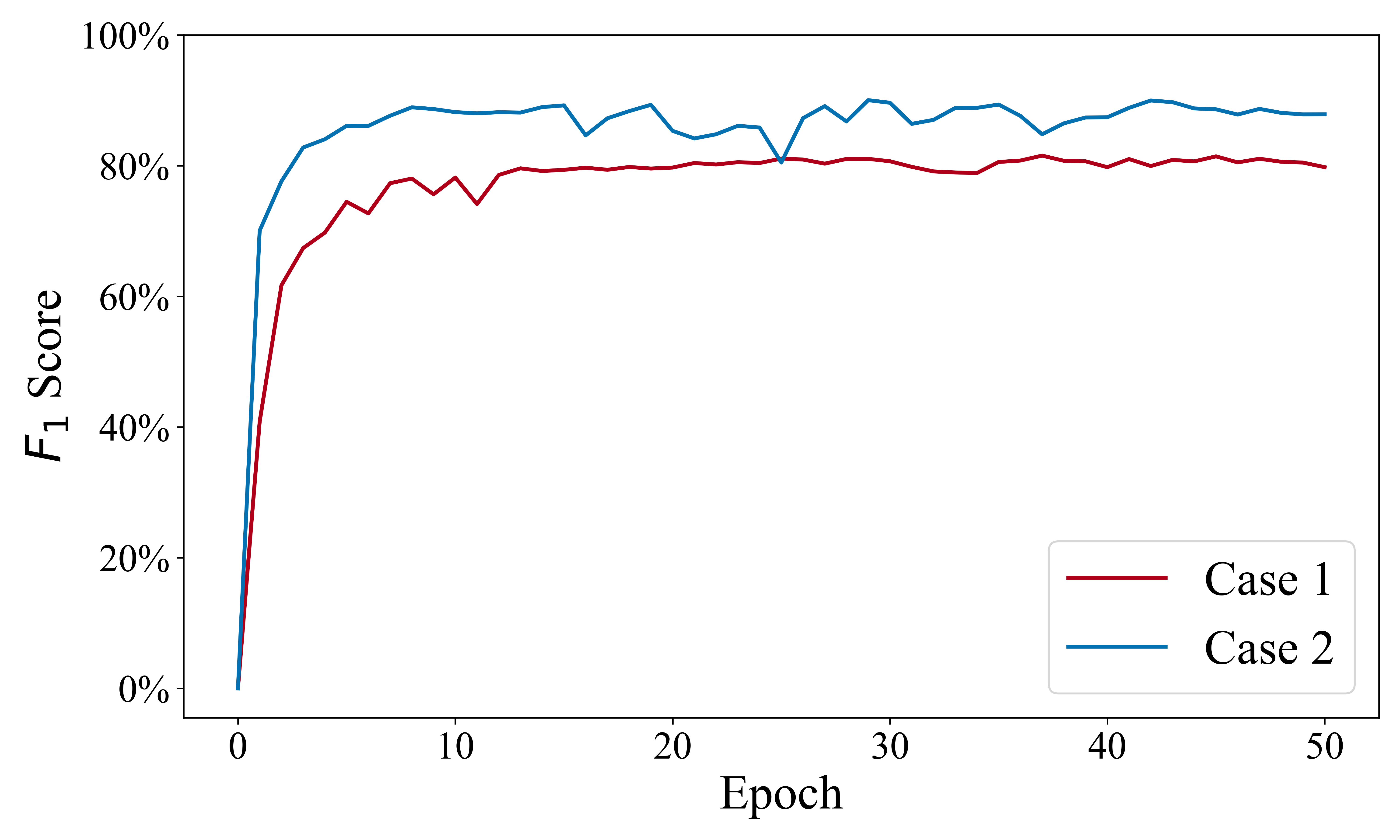} 
        \caption{Overall $F_1$ score}
        \label{fig:ft_ssl_case1}
    \end{subfigure}%
    \begin{subfigure}{.5\textwidth}
        \centering
        \includegraphics[width=\linewidth]{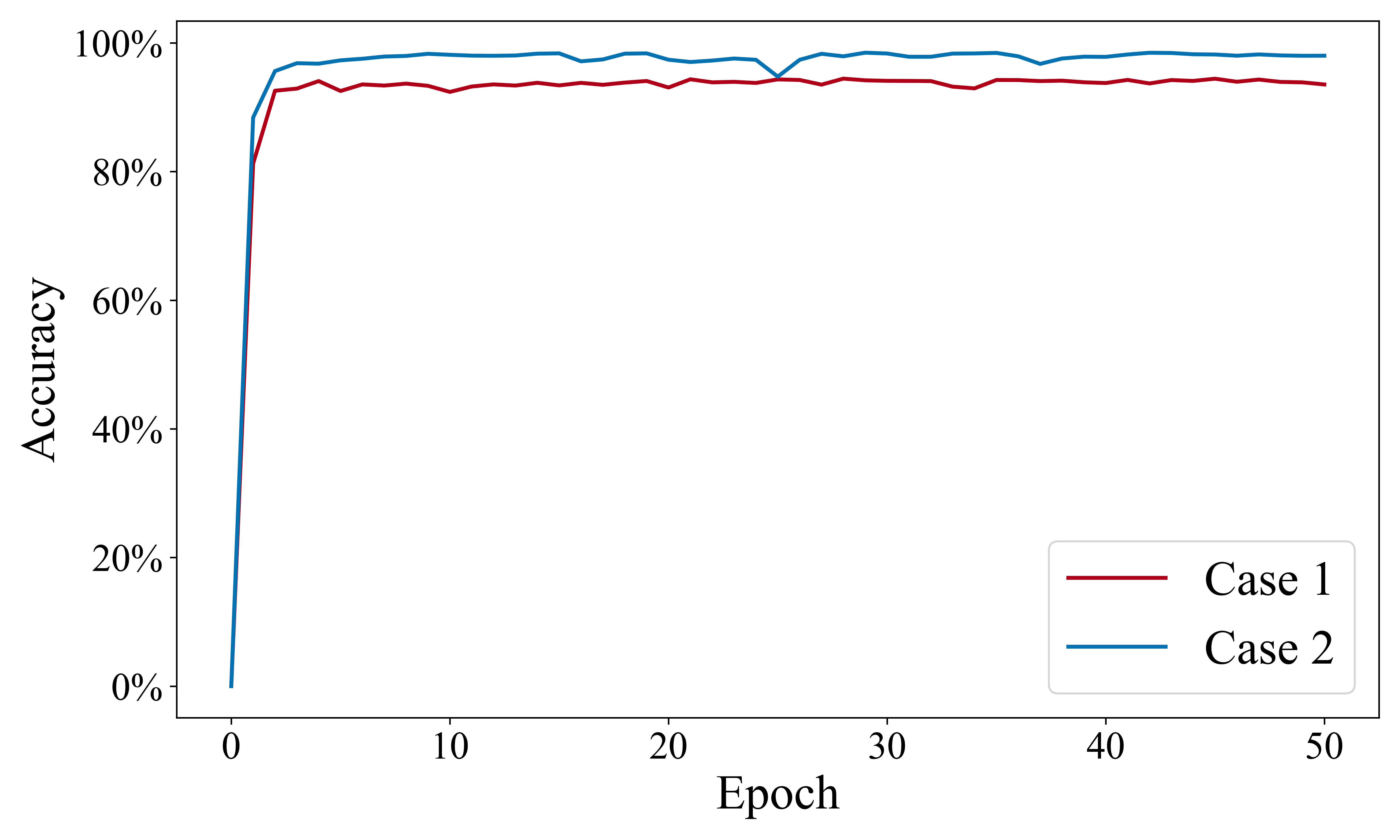} 
        \caption{Overall accuracy}
        \label{fig:ft_ssl_case2}
    \end{subfigure}
    \caption{Analysis of the performance in the fine-tuning stage.}
    \label{fig:fine-tune}
\end{figure*}

After pre-training,
Figure~\ref{fig:fine-tune} displays the $F_1$ score and accuracy of AE method during the fine-tuning stage.
In both Case 1 and Case 2, the $F_1$ score and accuracy increase rapidly, reaching optimal performance within approximately 10 epochs.
This demonstrates the efficiency of the fine-tuning process in leveraging the pre-trained encoder to achieve high performance in anomaly detection.

\subsubsection{Discussion}
The experimental results verify the effectiveness of SSL pre-training for anomaly detection in SHM data.
However, contrastive (SimCLR and Mixup) and generative-contrastive (GAN) SSL approaches generally underperform the generative (AE) method and, in some cases, even perform worse than purely supervised learning (SUP). 
In other data modalities, such as images and text, contrastive SSL methods have been reported to outperform generative SSL methods, achieving state-of-the-art performance on benchmark datasets~\cite{qiu2020pre,chen2020simple,goyal2021self}.
In the context of SHM data, however, AE emerges as the most effective and consistent pre-training method for anomaly detection.
This is likely due to the highly imbalanced data distribution during the pre-training stage, where the imbalance ratio between the majority and minority classes exceeds $10:1$.
Contrastive and generative-contrastive SSL methods suffer severely with such imbalanced data~\cite{assran2022hidden,balestriero2023cookbook}, often failing to learn the most discriminative features.

Furthermore, the overall performance is also influenced by the model's performance on minority abnormal patterns like `outlier' and `biased', which remains a large gap to improve.
This limitation also stems from the inherent imbalanced (long-tailed) distribution of real-world SHM data, where the `normal' pattern contains many data samples while minority patterns have fewer or rare instances. 
The majority of patterns dominate the training process, while rare types could be neglected.
Enhancing model's performance on minority abnormal data will be investigated in future research. 
Additionally, we observe that the model's performance of the current methods can be readily influenced by the data distribution in labeled datasets during fine-tuning. Designing the optimal configuration for the labeled dataset remains an open problem.

\label{}

\section{Conclusion}
In this work, we introduce emerging self-supervised learning (SSL) techniques in SHM data anomaly detection, which enable pre-training without the need for manually annotated labels.
Following pre-training, the SSL-based framework utilizes only a minimal amount of (hundreds of) labeled data for fine-tuning, demonstrating superior performance compared to purely supervised training.
The SSL-based framework is validated on SHM data from two in-service bridges with different structural types and abnormal patterns. 
The results show that autoencoder (AE) method is the most effective and consistent among the evaluated approaches, outperforming both purely supervised training and other SSL methods. 
However, while AE performs well on `normal` data and the majority of abnormal patterns, its performance on minority patterns shows a significant gap that needs to be improved.
Future research will focus on self-supervised learning with imbalanced (long-tailed) data, aiming to improve the detection of minority abnormal patterns.
Overall, SSL pre-training provides a practical solution for more automatic and low-labor SHM data cleansing. With very little human labeling process, it offers a new alternative for identifying and cleaning various abnormal patterns within the large-scale SHM data.

\section*{Data availability}
Data and code will be shared via a GitHub repository following the potential publication of this manuscript.

\section*{Acknowledgement}
The authors wish to express their gratitude for the financial support
received from the National Natural Science Foundation of China (51978508 \& 52278313), and the Top Discipline Plan of Shanghai Universities-Class I. This research is partly supported by the National Research Foundation, Prime Minister’s Office, Singapore under its Campus for Research Excellence and Technological Enterprise (CREATE) program, and the
Guangzhou-HKUST(GZ) Joint Funding Grant (No. 2023A03J0105), and the Guangdong Provincial Key Lab of Integrated Communication, Sensing and Computation for Ubiquitous Internet of Things (No.2023B1212010007).

\bibliographystyle{elsarticle-num} 
\biboptions{sort&compress}
\bibliography{local}

\end{document}